\documentclass[letterpaper]{article} 
\usepackage{aaai2026}  
\usepackage{times}  
\usepackage{helvet}  
\usepackage{courier}  
\usepackage[hyphens]{url}  
\usepackage{graphicx} 
\usepackage{amsmath} 
\usepackage{amsfonts} 
\usepackage{natbib}  
\usepackage{caption} 

\usepackage{algorithm}
\usepackage{algorithmic}
\usepackage{multicol}
\usepackage{multirow}
\usepackage{booktabs}
\usepackage{subcaption}

\usepackage{newfloat}
\usepackage{listings}
\DeclareCaptionStyle{ruled}{labelfont=normalfont,labelsep=colon,strut=off} 
\floatstyle{ruled}
\newfloat{listing}{tb}{lst}{}
\nocopyright

\usepackage{xcolor}

\title{BREPS: Bounding-Box Robustness Evaluation of Promptable Segmentation}
\author{
    Andrey Moskalenko\textsuperscript{\rm 1,2,3,4}\equalcontrib,
    Danil Kuznetsov\textsuperscript{\rm 3}\equalcontrib,
    Irina Dudko\textsuperscript{\rm 3},
    Anastasiia Iasakova\textsuperscript{\rm 3},\\
    Nikita Boldyrev\textsuperscript{\rm 2},
    Denis Shepelev\textsuperscript{\rm 2,3},
    Andrei Spiridonov\textsuperscript{\rm 2},\\
    Andrey Kuznetsov\textsuperscript{\rm 2},
    Vlad Shakhuro\textsuperscript{\rm 1,2,3}
}
\affiliations{
    \textsuperscript{\rm 1}Lomonosov Moscow State University\\
    \textsuperscript{\rm 2}FusionBrain Lab\\
    \textsuperscript{\rm 3}NUST MISIS\\
    \textsuperscript{\rm 4}IAI MSU\\
    and.v.moskalenko@gmail.com, kuznetsov.danil@proton.me\
}

\begin{document}

\maketitle

\begin{abstract}

Promptable segmentation models such as SAM have established a powerful paradigm, enabling strong generalization to unseen objects and domains with minimal user input, including points, bounding boxes, and text prompts.
Among these, bounding boxes stand out as particularly effective, often outperforming points while significantly reducing annotation costs.
However, current training and evaluation protocols typically rely on synthetic prompts generated through simple heuristics, offering limited insight into real-world robustness.
In this paper, we investigate the robustness of promptable segmentation models to natural variations in bounding box prompts.
First, we conduct a controlled user study and collect thousands of real bounding box annotations. 
Our analysis reveals substantial variability in segmentation quality across users for the same model and instance, indicating that SAM-like models are highly sensitive to natural prompt noise.
Then, since exhaustive testing of all possible user inputs is computationally prohibitive, we reformulate robustness evaluation as a white-box optimization problem over the bounding box prompt space.
We introduce BREPS, a method for generating adversarial bounding boxes that minimize or maximize segmentation error while adhering to naturalness constraints.
Finally, we benchmark state-of-the-art models across 10 datasets, spanning everyday scenes to medical imaging.

\end{abstract}

\begin{links}
    \link{Code}{https://github.com/emb-ai/BREPS}.
\end{links}

\begin{figure}[h!]
    \centering
    \includegraphics[width=1.0\columnwidth]{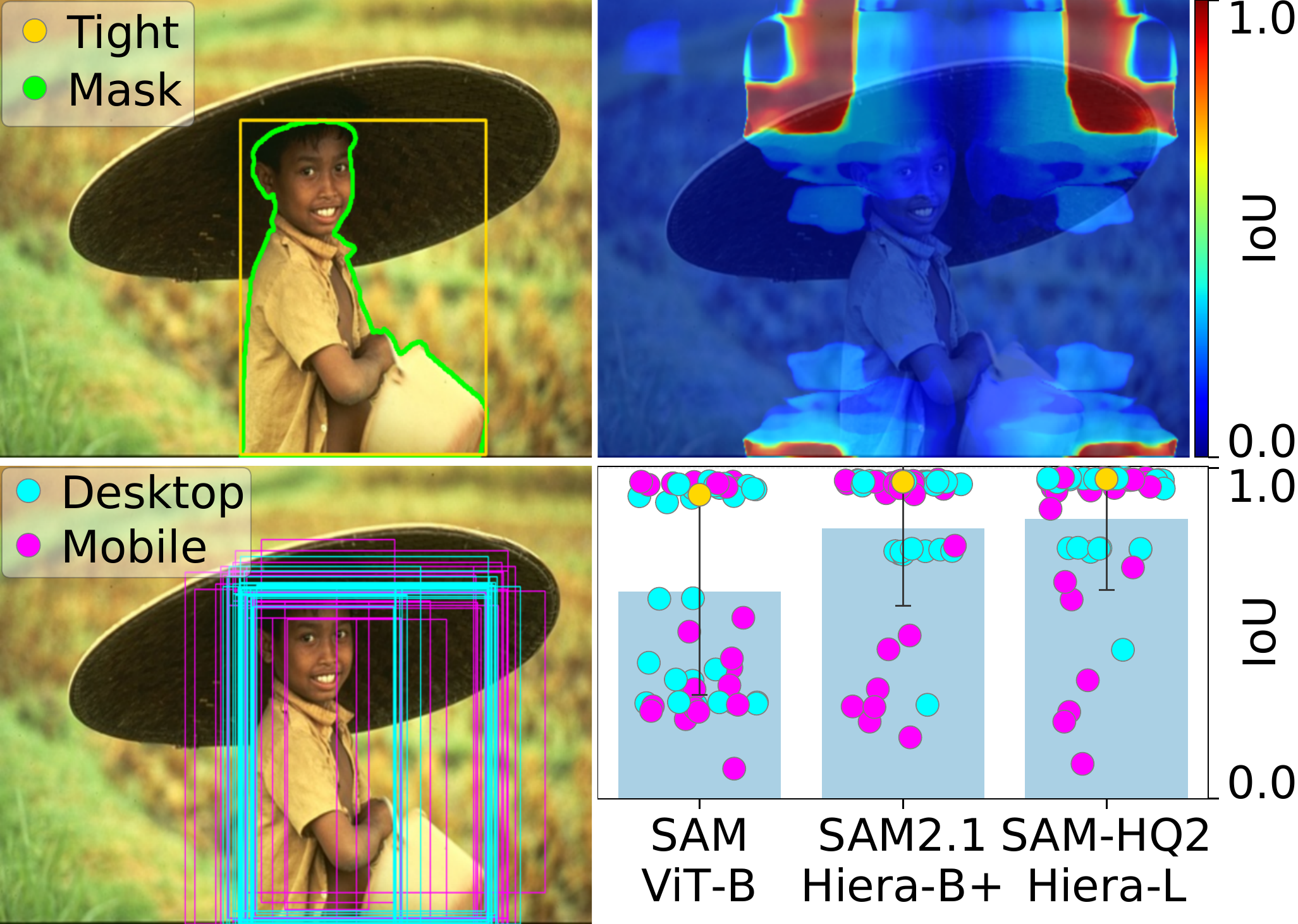}
    \caption{
    Top-left: image with ground-truth mask (green) and tight bbox (yellow).
    Bottom-left: real-users bboxes — desktop (cyan) and mobile (magenta).
    Top-right: IoU heatmap for SAM ViT-B; pixels shows IoU for a bbox anchored at that pixel and centered on the object.
    Bottom-right: IoU spread across 3 SOTA models for different users. We observed the large variability in IoU between individuals.
}
    \label{fig:clickbait}
\end{figure}

\begin{figure*}[h!]
    \centering
    \includegraphics[width=1.0\textwidth]{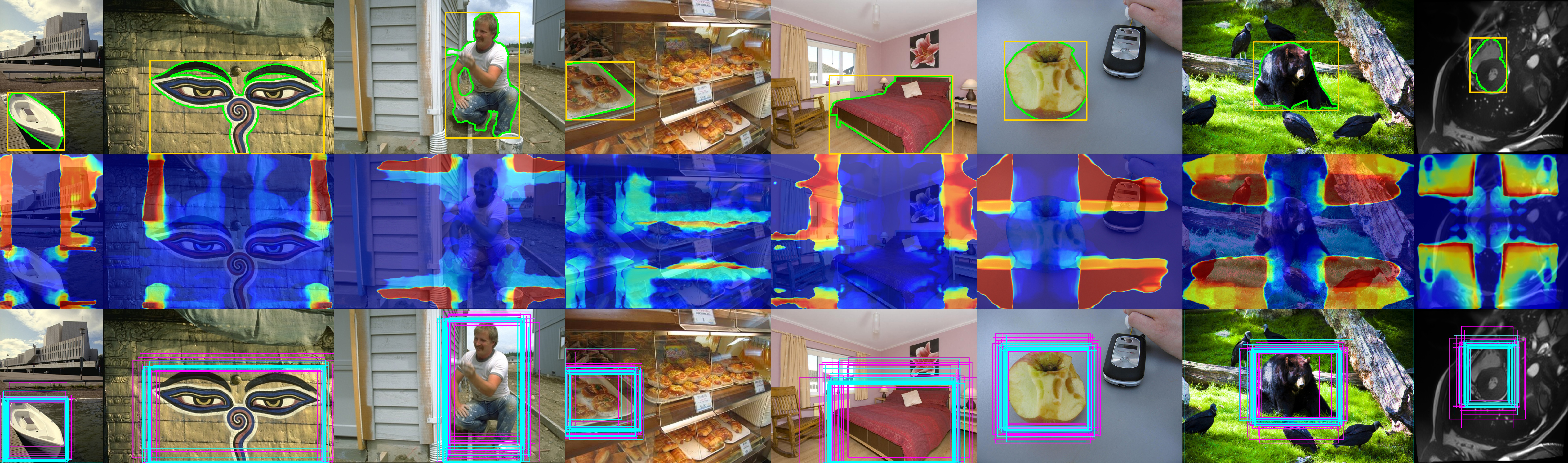}
    \caption{Top row: Berkeley, ADE20K, COCO, ACDC images with green mask and yellow tight bbox.
Middle: SAM ViT-B IoU heatmaps—pixel color shows IoU (0 blue → 1 red) for a bbox cornered at that pixel and centred on the object.
Bottom: user-drawn bboxes—desktop (cyan) vs mobile (magenta). Desktop prompts are tighter; user boxes vary and diverge from the tight bbox. Heatmaps reveal steep IoU drops from even 1-pixel shifts --- examples provided in Supplementary. Zoom for details.}
    \label{fig:heatmaps}
\end{figure*}

\section{Introduction}

Promptable segmentation has rapidly transitioned from a niche task to a fundamental problem. Its rise coincides with the emergence of foundational Segment Anything~\cite{kirillov2023segment} (SAM) model and successors. These models can output pixel-accurate masks of objects referred by a simple user prompt, e.g. point, bounding box, text, or a coarse mask. Nowadays promptable segmentation models are widely adopted to downstream applications ranging from photo and video editing to semi-automatic data labeling (CVAT~\cite{cvat}) and perception for robotics.

Among the available prompts, a bounding-box (bbox) is the most informative. Models that use bboxes usually output the best first-round masks~\cite{sam2, mazurowski2023segment}. These masks may be further improved with corrective prompts, i.e. points.
Almost all existing training and evaluation protocols use a simple bounding box sampling procedure. They use \textit{tight bbox} (e.g. bbox which is obtained from the boundaries of the instance on ground-truth segmentation mask) with a small jitter. This procedure doesn't include any prior information about how people actually draw bboxes. However, these models will ultimately be used by real users, who may encounter inconsistent model quality.

To gain insights on how models perform on real-users prompts, we conducted a large-scale user study with the help of 2,500 annotators.
Surprisingly, we found that — even though human boxes cluster closely, resulting mask quality (IoU) varies significantly between annotators (see Fig.~\ref{fig:clickbait}). We believe this instability of the models is due to the overfitting to synthetic bounding boxes and causes sim-to-real gap when we employ these models in real-world applications.

\begin{figure*}[h!]
    \centering
    \includegraphics[width=1.0\textwidth]{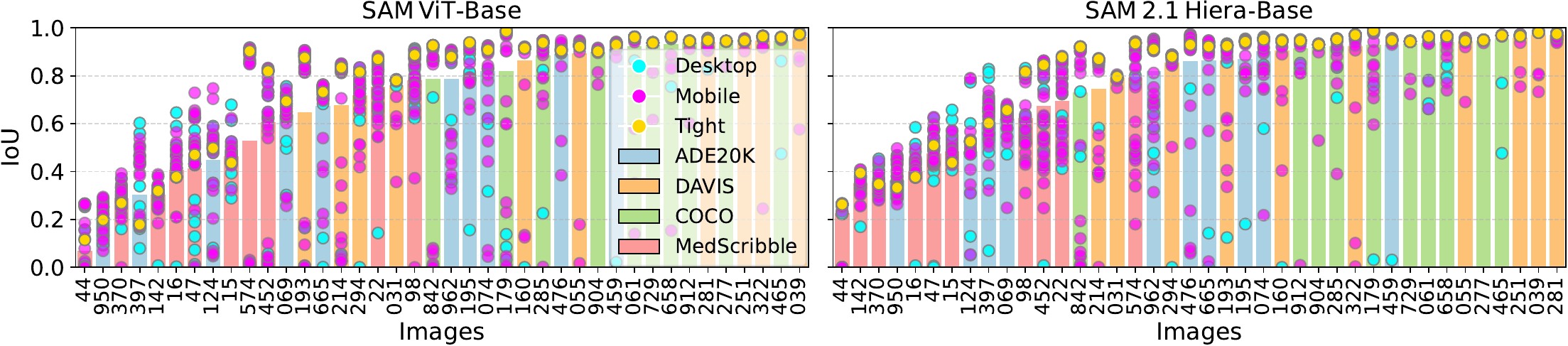}
    \caption{IoU spread for 10 random instances (3 general + 1 medical datasets). Points: desktop (cyan), mobile (magenta), tight bbox (yellow). Columns sorted by mean IoU over 50 users. SAM and SAM2.1 show large inter-user variance, while \textit{tight bboxes} consistently over-estimate quality.}
    \label{fig:user_bboxes}
\end{figure*}

Due to the limited availability of real-users prompts, we conducted a large-scale exhaustive search over valid bounding-box prompts for the most popular foundation models in both general-purpose and medical segmentation. 
We found that shifting the bounding box by just one pixel can change the quality significantly (see heatmaps in Fig.~\ref{fig:clickbait}, Fig.~\ref{fig:user_bboxes}).
However, exhaustively searching over every plausible bounding box is computationally prohibitive. We therefore recast the problem as a white-box adversarial attack on the bounding box prompt space.
A naive coordinate attack may collapse the box to a degenerate point; we prevent such trivial behavior using a regularizer derived from the empirical distributions obtained from our user study. 

Finally, we distill these ideas into a new robustness metric and perform the first comprehensive evaluation of 15 promptable segmentation models across 10 public datasets from general to medical domain.

Overall, our main contributions are as follows:
\begin{itemize}
    \item We conduct a pioneering controlled real-users study, collecting thousands of bounding box prompts across desktop and mobile settings. It reveals that users draw boxes that are far from the \textit{tight bboxes} and that state-of-the-art promptable segmentation models exhibit significant variability in performance across users.

    \item We introduce BREPS attack, a white-box optimization method for generating adversarial bounding boxes, guided by differentiable naturalness constraints to ensure realistic prompt perturbations.

    \item We perform a large-scale evaluation of state-of-the-art segmentation models across 10 datasets, uncovering average performance gaps of 30\% IoU under realistic prompt variation.
\end{itemize}

\noindent We believe that our methodology paves the way for promptable segmentation models that are more robust and higher-quality in real-world applications.

\section{Related Work}
\subsection{Promptable Segmentation}
\label{sec:promptable}

Promptable segmentation task is a generalization of the interactive segmentation problem, where the user guides the model with positive and negative visual inputs, typically clicks, to segment accurately the desired object \cite{ritm, focalclick, simpleclick, gaussian}. Promptable segmentation models form a new paradigm in which a single, versatile network can be adapted to various tasks at inference time by user \textit{prompts} — points, bounding boxes, text, or coarse masks. 

The SAM~\cite{kirillov2023segment} utilizes this idea by showing that a vision transformer~\cite{dosovitskiy2020vit}, trained once on a massive, automatically generated dataset, can generalize to novel objects, imaging modalities, and tasks with only a handful of user interactions. Since then, many SAM-like models have emerged, spanning general-domain~\cite{sam2,ke2023segment}, specialized medical ~\cite{medsam,sam-med2d}, efficiency-oriented ~\cite{mobilesam,mobilesamv2}, robustness-enhanced~\cite{robustsam, ppsam} models, and extensions that link grounding or text prompts to segmentation~\cite{groundedsam, groundeddino}.  

Among the rich prompt vocabulary, bounding boxes consistently yield the highest first-shot mask quality, often matching or even surpassing several corrective points in mean IoU~\cite{sam2, mazurowski2023segment}. 

\subsection{Adversarial Robustness}
\label{sec:adversarial}

Current benchmarks emphasize generalization across images and classes, but rarely across the prompt space itself. For example, previous works mostly sample bboxes from simple distributions (e.g. the \textit{tight bbox} with a small jitter), omitting the question of how real-users draw boxes and how sensitive models are to plausible variations. As a result, a model may achieve a strong average IoU under a single canonical prompt while exhibiting large performance variance under equally reasonable real-world alternatives.

Conventional adversarial attacks~\cite{goodfellow2014explaining} focus on perturbing the image. While promptable segmenters excel at generalizing across visual domains, some works have revealed that they are unstable to adversarial attacks~\cite{wang2024empirical}. \cite{segmentnothing} demonstrated that a geometry‐invariant pattern pasted onto any input can cause SAM to output empty masks, regardless of the user prompt. DarkSAM~\cite{zhou2024darksam} uses additive pixel noise to force the model to predict the background for any prompt. BadSAM~\cite{Guan_Hu_Zhou_Zhang_Li_Liu_2024} implements a finetuning procedure such that a specific trigger prompt yields a preselected mask. RoBox-SAM~\cite{huang2024robust}, PP-SAM~\cite{ppsam} and ASAM~\cite{li2024asam} tuning the model with randomly perturbed prompts so it remains accurate under box or point jitter. Yet they assume small, synthetic perturbations and therefore fail to cover real-users behavior as in our study.

TETRIS~\cite{moskalenko2024tetris} probe the robustness of interactive-segmentation models to click prompts by using gradient-based optimization. Unfortunately, the resulting points are not realistic from a human-annotator standpoint — for example, they can land precisely on object boundaries, a behavior real-users rarely exhibit. We address this gap by introducing a regularization that explicitly constrains the search to \textit{human-plausible} bounding box prompts.

The authors of RClicks~\cite{antonov2024rclicks} train a clickability model on a corpus of genuine user clicks and leverage it to benchmark the accuracy of click-driven models. However, their evaluation is limited to a black-box setup with random sampling, and it entirely omits the widely used — and often higher-performing bounding box prompt modality.

\subsection{Segmentation Prompts User Studies}
\label{sec:userstudies}

Early efforts to understand how people annotate images at scale focused on crowdsourcing visual object labels. \cite{su2012crowdsourcing} quantified how quickly non-experts could draw \textit{tight bboxes}. Their analysis uncovered systematic biases — e.g. small objects are often missed — that later motivated tighter interaction loops. 

Building on this, a line of work examined how much human effort each supervision modality actually costs. \cite{papadopoulos2017extreme} introduced Extreme Clicking — four corner clicks instead of a full box — to cut annotation time $\sim$7 s per object without hurting IoU. Click’n’Cut~\cite{clickncut} and DAVIS interactive video segmentation tracks~\cite{Pont-Tuset_arXiv_2017} extended the paradigm to pixel-level masks, demonstrating that several iterative rounds of semi-supervised segmentation with user feedback can match the quality of fully-supervised segmentation.

\begin{table*}[h!]
\centering
\fontsize{10pt}{13pt}\selectfont
\tabcolsep=3pt
\begin{tabular}{c|c|ccccccc}
\toprule
Method & Backbone & \multicolumn{1}{c}{GrabCut} & \multicolumn{1}{c}{Berkeley} & \multicolumn{1}{c}{DAVIS} & \multicolumn{1}{c}{COCO} & \multicolumn{1}{c}{TETRIS} & \multicolumn{1}{c}{PASCAL} & \multicolumn{1}{c}{ADE20K} \\ \hline
MobileSAM & ViT\texttt{-}T & 88.92±11.49 & 83.67±10.73 & 76.54±10.21 & 84.46±10.27 & 75.65±11.56 & 82.05±\phantom{0}9.48 & \underline{71.20}±13.83 \\ \hline
\multirow{3}{*}{SAM} & ViT\texttt{-}B & 86.20±16.87 & 81.32±14.31 & 80.23±13.78 & 80.83±15.98 & 70.99±16.13 & 82.41±13.44 & 62.17±15.73 \\
& ViT\texttt{-}L & 88.59±13.45 & 84.50±12.85 & 83.85±12.36 & 85.16±13.80 & 79.57±14.46 & 86.15±10.69 & 64.41±14.55 \\
& ViT\texttt{-}H & 89.11±13.48 & 83.87±12.38 & 83.92±12.28 & 85.14±12.63 & 79.45±14.82 & 86.22±\phantom{0}9.98 & 63.68±13.02 \\ \hline
\multirow{3}{*}{SAM\texttt{-}HQ} & ViT\texttt{-}B & 91.74±\phantom{0}9.00 & \underline{88.29}±\phantom{0}9.69 & 83.92±\phantom{0}9.73 & 86.56±\phantom{0}9.53 & 80.80±11.10 & 85.99±\phantom{0}8.13 & 68.37±14.63 \\
& ViT\texttt{-}L & 92.15±\phantom{0}8.74 & 88.09±10.56 & 86.12±\phantom{0}9.92 & \textbf{87.33}±10.21 & \textbf{83.20}±11.91 & \textbf{87.97}±\phantom{0}8.26 & \textbf{71.24}±14.68 \\
& ViT\texttt{-}H & \underline{92.30}±\phantom{0}9.01 & \textbf{88.44}±\phantom{0}9.63 & \textbf{86.28}±\phantom{0}9.96 & \underline{86.90}±10.28 & \underline{83.16}±11.99 & \underline{87.81}±\phantom{0}8.06 & 68.68±13.83 \\ \hline
\multirow{4}{*}{SAM 2.1} & Hiera\texttt{-}T & 90.09±12.97 & 85.31±12.78 & 83.65±11.73 & 84.36±12.34 & 75.87±16.33 & 86.79±\phantom{0}8.99 & 68.56±14.56 \\
& Hiera\texttt{-}S & 90.36±13.06 & 86.51±12.64 & 82.57±14.25 & 84.60±13.47 & 77.14±16.00 & 85.82±11.05 & 66.44±16.07 \\
& Hiera\texttt{-}B$^{+}$ & 89.72±14.12 & 84.90±14.07 & 83.39±15.59 & 84.80±14.11 & 78.38±16.81 & 87.18±11.27 & 68.40±15.39 \\
& Hiera\texttt{-}L & 90.31±13.18 & 85.82±12.91 & 85.67±11.97 & 85.45±13.12 & 78.74±15.36 & 87.26±10.60 & 65.36±15.63 \\ \hline
SAM\texttt{-}HQ 2 & Hiera\texttt{-}L & \textbf{92.35}±11.22 & 88.19±12.67 & \underline{86.25}±11.76 & 86.59±13.31 & 81.71±15.95 & 87.47±10.55 & 69.67±17.20 \\ \hline
\multirow{3}{*}{RobustSAM} & ViT\texttt{-}B & 78.39±\phantom{0}9.63 & 74.65±10.75 & 60.36±10.98 & 75.74±\phantom{0}9.44 & 62.06±\phantom{0}9.87 & 73.10±\phantom{0}8.56 & 54.29±10.58 \\
& ViT\texttt{-}L & 64.75±11.17 & 45.16±10.03 & 27.97±\phantom{0}7.41 & 52.70±10.15 & 42.07±\phantom{0}9.48 & 56.08±10.11 & 32.82±\phantom{0}9.91 \\
& ViT\texttt{-}H & 64.82±\phantom{0}6.40 & 49.30±\phantom{0}5.60 & 27.08±\phantom{0}2.99 & 61.54±\phantom{0}5.88 & 53.13±\phantom{0}6.12 & 58.90±\phantom{0}5.54 & 52.40±\phantom{0}6.79 \\ \bottomrule
\end{tabular}
\caption{Promptable segmentation models IoU performance on real-users bounding boxes on general segmentation datasets. Best results are in \textbf{bold}, the second best is \underline{underlined}. The standard deviation was computed across 50 users and averaged over all images in the datasets. Please refer to the Supplementary for separated desktop/mobile devices results.}
\label{tab:general_users}
\end{table*}
 
Several studies have analyzed real-users behaviour in interactive segmentation. \cite{Myers-Dean_2024_WACV} recorded free interactions and found that circling was preferred by users in more than 70\% of cases, while less than 15\% for clicks. Bboxes can be extracted from circling and put into the promptable segmentation models. RClicks~\cite{antonov2024rclicks} crowdsourced corrective clicks and trained a clickability model to emulate human click distributions.

Despite advances, bounding box prompts remain underexplored. Existing analyses either treat boxes as noise-free rectangles or perturb them with small Gaussian jitter~\cite{ppsam}, failing to capture the plausible yet “unlucky’’ boxes we observe in practice. In contrast to prior studies, we provide the first systematic look at bbox realism, variability, and adversarial vulnerability, laying the groundwork for robustness-aware training and evaluation protocols.

\section{Real-Users Study}
To investigate user behaviour when drawing bounding boxes we conducted a large‑scale crowdsourcing experiment with 2,500 participants.

\subsection{Data Selection}
\label{datasets}

For this study, we required datasets that provide both reference images and ground‑truth segmentation masks.  
To guarantee coverage across domains and use cases, we mixed general‑purpose and medical segmentation sets.  
We sampled 50 instances from each of the popular promptable general and medical datasets (total \# of instances after the dash):
\begin{itemize}
\item GrabCut --- 50~\citep{rother2004grabcut};
\item Berkeley --- 100~\citep{arbelaez2011contour};
\item DAVIS --- 345~\citep{perazzi2016davis}, subset from \cite{ritm};
\item COCO-MVal --- 800~\citep{lin2014coco}, subset from \cite{ritm};
\item TETRIS --- 2,531~\citep{moskalenko2024tetris};
\item ADE20K --- 707,868~\citep{zhou2017ade20k};
\item PASCAL-VOC2012 --- 19,694~\citep{everingham2010pascal};
\item ACDC --- 100~\citep{bernard2018acdc};
\item BUID --- 780~\citep{aldhabyani2020buid};
\item MedScribble --- 56~\cite{wong2024scribbleprompt}, 3--5 available samples per multiple medical datasets in the split.
\end{itemize}

Overall, from the 10 datasets we selected 500 images. To minimize load time, every image shown to workers was down‑scaled so that its longer side did not exceed 1024px.

\subsection{Crowdsourcing Setup}

The design of the user-study begins with the choice of what can be considered a bbox and the method of display. The difference is that, unlike the modality of a point, which is made by a regular click, a bbox can be drawn in several ways while obtaining the same bbox by coordinates:

\begin{itemize}
    \item  The classic drag‑and‑drop from the top‑left to bottom‑right corner as used by CVAT~\cite{cvat}, Label Studio~\cite{labelstudio}, and LabelImg~\cite{labelimg};
    \item Two‑click schemes recording the two opposite corners (two-point mode in CVAT~\cite{cvat});
    \item Polygons or lassos that are automatically converted to bboxes \cite{v7darwin}, \cite{supervisely};
    \item Extreme clicking, i.e. four extreme points on the object contour~\cite{papadopoulos2017extreme}.
\end{itemize}

Since drag‑and‑drop bbox labeling is the most widespread in current annotation pipelines, we adopted it in our study.

\begin{figure}[ht!]
    \centering
    \includegraphics[width=1.0\columnwidth]{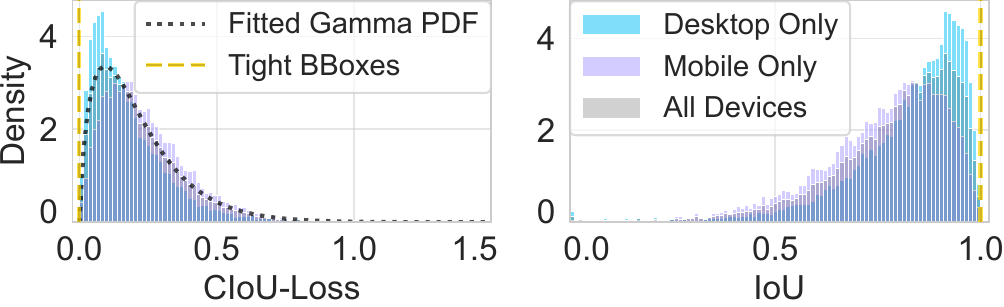}
    \caption{CIoU-Loss and IoU density plots for real-users drawn vs. \textit{tight bboxes}. Black dashed shows fitted Gamma PDF. Mobile prompts (magenta) skew to lower overlap --- we observe that on mobile devices, bboxes are more deviated from the \textit{tight bboxes} than on desktop devices.}
    \label{fig:bboxes_quality}
\end{figure}

\paragraph{Display Bias Elimination}

Directly displaying the ground‑truth mask would predispose participants and introduce anchoring bias~\cite{draws2021checklist}. Our bbox drawing task, therefore, had to feel as natural as if users were selecting the object on their own.

Following the interface guidelines~\cite{bauchwitz2025task} and ~\cite{antonov2024rclicks}, we implemented a three‑stage protocol in our real-users study:
\begin{enumerate}
  \item \textbf{Free viewing} — the raw image is shown for 3s; bbox drawing is disabled.
  \item \textbf{Target memorization} — only the target region is visible for another 3s, the rest being grayed out; bbox drawing remains disabled.
  \item \textbf{Annotation} — the mask disappears, the full image reappears, and the worker may draw exactly one bbox.
  \item \textbf{Repetition} — if the assessor does not remember the target area mask, he can go through all three steps from the beginning by pressing the corresponding button.
\end{enumerate}

\begin{table}[h!]
\centering
\fontsize{9pt}{11.8pt}\selectfont
\tabcolsep=1.9pt
\begin{tabular}{c|c|ccc}
\toprule
Method & Backbone & \multicolumn{1}{c}{ACDC} & \multicolumn{1}{c}{BUID} & \multicolumn{1}{c}{MedScribble} \\ \hline
MobileSAM & ViT\texttt{-}T & 79.29±\phantom{0}9.24 & 56.64±10.21 & 46.88±\phantom{0}8.54 \\ \hline
\multirow{3}{*}{SAM} & ViT\texttt{-}B & 83.43±10.10 & 61.19±\phantom{0}9.78 & 48.82±12.79 \\
 & ViT\texttt{-}L & \textbf{84.26}±\phantom{0}9.83 & 63.35±\phantom{0}9.06 & \underline{54.22}±\phantom{0}8.97 \\
 & ViT\texttt{-}H & 83.09±\phantom{0}9.38 & 63.12±\phantom{0}9.26 & 53.73±\phantom{0}9.86 \\ \hline
\multirow{3}{*}{SAM\texttt{-}HQ} & ViT\texttt{-}B & 83.08±\phantom{0}8.52 & 58.39±\phantom{0}9.81 & 49.86±10.21 \\
 & ViT\texttt{-}L & \underline{83.55}±\phantom{0}9.08 & 63.40±\phantom{0}8.48 & 53.10±\phantom{0}8.31 \\
 & ViT\texttt{-}H & 82.83±\phantom{0}8.68 & 63.02±\phantom{0}8.90 & 53.35±\phantom{0}8.65 \\ \hline
\multirow{4}{*}{SAM 2.1} & Hiera\texttt{-}T & 80.21±10.87 & 60.15±12.56 & \textbf{54.69}±10.20 \\
 & Hiera\texttt{-}S & 78.87±10.16 & 63.20±10.32 & 54.10±11.66 \\
 & Hiera\texttt{-}B$^{+}$ & 80.34±11.11 & 61.27±12.24 & 51.85±11.00 \\
 & Hiera\texttt{-}L & 79.11±10.52 & \underline{68.11}±\phantom{0}9.56 & 54.19±12.27 \\ \hline
SAM\texttt{-}HQ 2 & Hiera\texttt{-}L & 79.54±10.06 & \textbf{69.25}±10.19 & 53.13±11.68 \\ \hline
\multirow{3}{*}{RobustSAM} & ViT\texttt{-}B & 62.20±10.44 & 34.87±\phantom{0}7.20 & 46.01±\phantom{0}8.22 \\
 & ViT\texttt{-}L & 43.09±\phantom{0}8.28 & 15.39±\phantom{0}4.46 & 27.63±\phantom{0}7.62 \\
 & ViT\texttt{-}H & 46.81±\phantom{0}6.15 & 47.25±\phantom{0}5.34 & 36.77±\phantom{0}4.42 \\ \bottomrule
\end{tabular}

\caption{Promptable segmentation models' performance on real-users bounding boxes on medical segmentation datasets. Best results are in \textbf{bold}, the second best is \underline{underlined}. The standard deviation was computed across 50 users and averaged over all images in the datasets.}
\label{tab:medical_users}
\end{table}

\noindent The annotation instruction for assessors:

\begin{itemize}
\item After opening the task, wait for the images to load within 30–60 seconds.
\item After clicking the \textit{Start viewing} button, you will be shown an image, then the area of interest on a gray background.
\item Your task is to select \textbf{one rectangle} covering the area designated in the previous step.
\end{itemize}

The interface ran on both desktops and mobile devices.
Each image was scaled to fill the available screen area; the bbox outline had a stroke width of $1\%$ of the shorter image side.  
To compensate for finger thickness on touch screens on a near-border instances, a $5\%$ from each side padding was added beyond the image borders. However, bbox drawings beyond borders results in coordinates clipping.

Each worker produced 10 bboxes for 10 different images. A hard time limit of 15 minutes was enforced, and the mean completion time was under 4 minutes. In all experiments, we use Toloka~\cite{toloka2025} crowdsourcing vendor.

\subsection{Real-Users Behavior Analysis}
\label{real_user_behaviour}

In total, we collected 25,000 bounding boxes from 2,500 people (50 boxes per image), with half of the annotations drawn on desktop devices and half on mobile devices. In this section we analyze the collected user inputs, set bbox quality metrics and propose differentiable bbox realism regularization, which will be utilized in our BREPS pipeline.

\subsubsection{Bounding-Box Quality Measuring}

We objectively measure the quality of observed bounding boxes w.r.t. the perfect \textit{tight bbox}. Thus, we use the Intersection over Union (IoU) and the CIoU-Loss~\cite{zheng2020distance} (Complete IoU), which is computed as follows:
$$
\mathcal{L}_{\mathrm{CIoU}}\left(B, B^*\right) = 1 - \operatorname{IoU}\left(B, B^*\right) + \frac{\rho^2\left(\mathbf{b}, \mathbf{b}^*\right)}{c^2} + \alpha v
$$

\noindent Where,
\begin{itemize}
  \item $
\operatorname{IoU}\left(B, B^*\right) =\frac{\lvert B \cap B^*\rvert}{\lvert B \cup B^* \rvert}
$
  \item $B$ and $B^{*}$ denote the observed and ground-truth \textit{tight bounding boxes}, obtained from segmentation masks.
  \item $\mathbf{b}=(x,y), \mathbf{b}^{*}=(x^{*},y^{*})$ are the centers of $B$ and $B^{*}$.
  \item $\rho(\mathbf{b},\mathbf{b}^{*})$ is the Euclidean distance between the centers.
  \item $c$ is the diagonal length of the smallest enclosing box covering both $B$ and $B^{*}$.
  \item $v$ measures the consistency of aspect ratios,
    \[
      v = \frac{4}{\pi^2}\Bigl(\arctan\!\tfrac{w^{*}}{h^{*}} - \arctan\!\tfrac{w}{h}\Bigr)^{2},
    \]
    where $(w,h), (w^{*},h^{*})$ are the $B, B^{*}$ width and height.
  \item $\alpha$ is a positive trade-off parameter defined as
    \[
      \alpha = \frac{v}{(1 - \mathrm{IoU}(B, B^{*})) + v}.
    \]
\end{itemize}

\subsubsection{Bounding-Box Quality Analysis}
The obtained distributions of the values of quality functionals are shown in Fig.~\ref{fig:bboxes_quality} As a ground-truth bbox, we took a \textit{tight bbox} obtained from a ground-truth segmentation mask. We measured the performance separately for mobile and desktop users. We observed that users on phones are statistically significantly worse ($U$-test~\cite{mann1947test} with $p_{value} < 0.01$) in IoU and CIoU-Loss, which we attribute to the complexity of drawing on a small screen and the lower precision of a finger compared to a mouse cursor.

\subsubsection{Measuring the Bounding-Box Realism}

We are interested in constructing a bounding‑box realism regularizer for our optimization procedure.  
Since the CIoU-Loss exhibits smoother properties and allows gradients to flow even when the boxes do not overlap, we have chosen to build on it.  
We fitted several candidate distributions to a combined empirical sample (itself drawn from Beta and Gamma distributions).  
The best fit was achieved with a Gamma distribution ($k = 1.789, \theta = 0.121$) over the CIoU-Loss values between a bbox $B$ and its \textit{tight bbox} $B^{\star}$. Using the probability density function (PDF) of Gamma, the realism of any bbox $B$ can be assessed directly through the log‑likelihood ($\star$):
\[
\begin{aligned}
\log \operatorname{PDF}(X; k,\theta) &=
  (k-1)\ln\!X - \tfrac{X}{\theta} - k\ln\theta - \ln\Gamma(k) \\[2pt]
X &= \mathcal{L}_{\mathrm{CIoU}}\!\bigl(B,B^*\bigr)
\end{aligned}
\]

Obtained histograms and fitted PDF illustrated in Fig.~\ref{fig:bboxes_quality}.

\begin{figure}[t!]
    \centering
    \includegraphics[width=1.0\columnwidth]{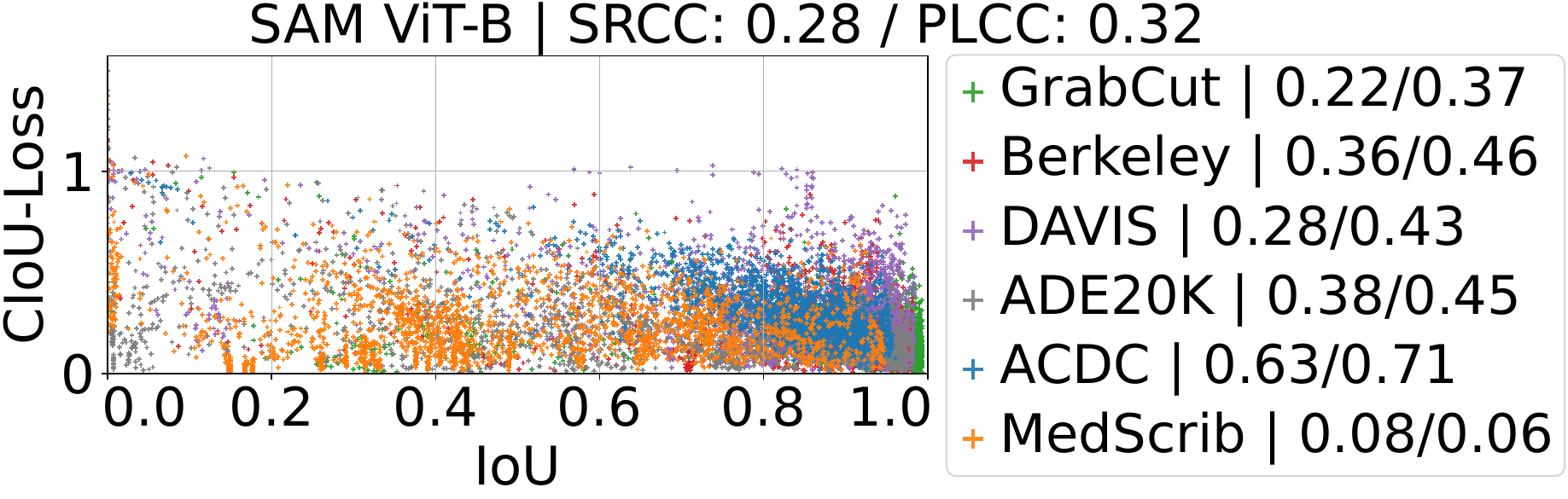}
    \caption{CIoU-Loss/IoU scatters for SAM ViT-B model; dataset labels show Spearman/Pearson correlations. We don't observe high correlations on 9 out of 10 datasets and associate the highest correlation of ACDC with the similarity and simplicity of instances in this dataset.}
    \label{fig:corrs}
\end{figure}

\subsection{Models Robustness on Real-Users}

We compared 15 promptable segmentation models checkpoints using collected real-users prompts (MobileSAM~\cite{mobilesam}, SAM~\cite{kirillov2023segment}, SAM-HQ, SAM-HQ2~\cite{ke2023segment}, SAM2~\cite{sam2}, RobustSAM~\cite{robustsam}. For these general segmentation models, we used all datasets from Sec.~\ref{datasets}. 

To compare models, we use the generally accepted IoU metric averaged across all instances in the dataset. Additionally, we present the standard deviation of model quality between different real-users. Results presented in Tab.~\ref{tab:general_users}, some examples provided in Fig.~\ref{fig:user_bboxes}.

\subsection{Bounding-Box Quality vs Model Performance}

We explored whether the quality of a bounding box (relative to the ideal \textit{tight box}) correlates with the quality of the resulting segmentation for that box. To do this, we compute Spearman and Pearson correlations for several models. A subset of the results for SAM is provided in Fig.~\ref{fig:corrs}.

We observed a weak correlation ($\le 0.4$ SRCC on average on most datasets), indicating a more complex relationship between bbox perturbations and the resulting segmentation quality. This confirms the need for an assessment of the stability of promptable models and not just an assessment of the quality of the prompts at the input.  Please refer to the Supplementary for more detailed analysis of collected bboxes.

\section{Proposed BREPS Attack}

\begin{table*}[ht]
\centering
\fontsize{9pt}{12pt}\selectfont
\tabcolsep=2.25pt
\begin{tabular}{c|c|cccccccccccccccccccc}
\toprule
\multirow{2}{*}{Method} & \multirow{2}{*}{Backbone} &  & \multicolumn{4}{c}{COCO-MVal} &  & \multicolumn{4}{c}{PASCAL-VOC2012} &  & \multicolumn{4}{c}{ADE20K} &  & \multicolumn{4}{c}{BUID} \\ \cline{4-7} \cline{9-12} \cline{14-17} \cline{19-22}
 &  &  & Tight & Min & Max & $\Delta$ &  & Tight & Min & Max & $\Delta$ &  & Tight & Min & Max & $\Delta$ &  & Tight & Min & Max & $\Delta$ \\ \cline{1-2} \cline{4-7} \cline{9-12} \cline{14-17} \cline{19-22}
MobileSAM            & ViT-T   &  & 86.88 & 71.50 & 89.40 & 17.90 &  & 85.81 & 70.00 & 87.81 & 17.81 &  & 79.62 & \textbf{49.66} & 83.68 & \underline{34.02} &  & 73.14 & 54.27 & 77.95 & 23.68 \\ \cline{1-2} \cline{4-7} \cline{9-12} \cline{14-17} \cline{19-22}
\multirow{3}{*}{SAM} & ViT-B      &  & 86.95 & 67.73 & 89.76 & 22.03 &  & 86.13 & 60.54 & 88.43 & 27.89 &  & \textbf{79.93} & 42.31 & 84.26 & 41.95 &  & 77.55 & 55.87 & 81.07 & 25.20 \\
                     & ViT-L      &  & 87.63 & 71.06 & 89.36 & 18.30 &  & \underline{88.21} & 71.26 & 89.40 & 18.14 &  & 79.43 & 39.20 & 83.48 & 44.28 &  & 75.39 & \textbf{58.99} & 78.73 & \underline{19.74} \\
                     & ViT-H      &  & 87.48 & \underline{74.72} & 89.03 & 14.31 &  & \textbf{88.46} & 74.85 & 89.56 & 14.70 &  & 77.97 & \underline{49.24} & 82.27 & \textbf{33.03} &  & 75.28 & \underline{58.86} & 79.05 & 20.19 \\ \cline{1-2} \cline{4-7} \cline{9-12} \cline{14-17} \cline{19-22}
\multirow{3}{*}{SAM-HQ} & ViT-B   &  & 87.66 & 70.67 & 90.36 & 19.69 &  & 86.06 & 73.15 & 88.66 & 15.51 &  & 76.79 & 32.70 & 84.21 & 51.51 &  & 75.95 & 55.53 & 80.21 & 24.67 \\
                       & ViT-L   &  & 87.77 & 74.68 & 89.90 & 15.22 &  & 87.92 & \underline{77.46} & 89.63 & \underline{12.17} &  & \underline{79.73} & 33.11 & \underline{85.25} & 52.15 &  & 75.28 & 58.52 & 79.52 & 21.00 \\
                       & ViT-H   &  & 87.67 & \textbf{76.73} & 89.52 & \textbf{12.79} &  & 88.14 & \textbf{79.89} & \underline{89.71} & \textbf{9.82} &  & 78.77 & 40.73 & 84.90 & 44.17 &  & 75.12 & 57.49 & 79.36 & 21.87 \\ \cline{1-2} \cline{4-7} \cline{9-12} \cline{14-17} \cline{19-22}
\multirow{4}{*}{SAM 2.1} & Hiera-T  &  & 86.87 & 55.80 & 89.76 & 33.96 &  & 87.21 & 46.36 & 88.55 & 42.19 &  & 77.69 & 22.72 & 82.95 & 60.23 &  & 77.63 & 43.90 & 82.23 & 38.34 \\
                         & Hiera-S  &  & 87.09 & 52.55 & 89.60 & 37.05 &  & 87.51 & 46.88 & 88.70 & 41.82 &  & 75.34 & 15.46 & 80.34 & 64.89 &  & 76.53 & 45.74 & 80.64 & 34.89 \\
                         & Hiera-B+ &  & \underline{88.31} & 57.52 & \underline{90.69} & 33.18 &  & 87.79 & 54.50 & 89.21 & 34.71 &  & 79.01 & 22.32 & 84.27 & 61.95 &  & \underline{80.47} & 47.01 & \underline{83.90} & 36.89 \\
                         & Hiera-L  &  & 87.80 & 59.77 & 90.08 & 30.30 &  & 88.32 & 56.18 & 89.68 & 33.50 &  & 75.87 & 23.39 & 81.73 & 58.34 &  & 80.22 & 49.04 & 83.19  & 34.15 \\ \cline{1-2} \cline{4-7} \cline{9-12} \cline{14-17} \cline{19-22}
SAM-HQ 2              & Hiera-L   &  & \textbf{89.46} & 66.60 & \textbf{91.70} & 25.10 &  & \underline{88.44} & 71.64 & \textbf{90.34} & 18.69 &  & 79.60 & 23.12 & \textbf{85.79} & 62.66 &  & \textbf{82.71} & 51.95 & \textbf{86.10} & 34.15 \\ \cline{1-2} \cline{4-7} \cline{9-12} \cline{14-17} \cline{19-22}
\multirow{3}{*}{RobustSAM} & ViT-B   &  & 77.66 & 31.52 & 82.57 & 51.05 &  & 76.73 & 31.46 & 79.74 & 48.28 &  & 72.44 & 10.40 & 78.13 & 67.73 &  & 67.50 & 25.75 & 71.33 & 45.58 \\
                           & ViT-L   &  & 52.06 & 4.22 & 57.92 & 53.70 &  & 68.40 & 5.05 & 73.92 & 68.86 &  & 53.85 & 1.86 & 61.31 & 59.45 &  & 29.16 & 2.29 & 34.77 & 32.48 \\
                           & ViT-H   &  & 34.15 & 23.37 & 37.49 & \underline{14.12} &  & 49.68 & 34.05 & 53.99 & 19.95 &  & 54.39 & 23.08 & 60.89 & 37.81 &  & 29.35 & 19.71 & 33.23 & \textbf{13.52} \\ \bottomrule
\end{tabular}
\caption{BREPS attack results on state-of-the-art promptable segmentation models. Results provided for three general segmentation datasets and one medical one, the remaining are provided in the Supplementary due to limited space. Best results are in \textbf{bold}, the second best is \underline{underlined}.}
\label{tab:breps_attack}
\end{table*}

\subsection{Exhaustive Search}

Since limited real‑user and sampled prompts cannot cover every possible configuration of bounding boxes, we perform a computationally expensive exhaustive search on several instances from each dataset.

However, if we fully parameterize a bounding box by its four parameters \((x_1, y_1, x_2, y_2)\) or \((x, y, h, w)\), an image of size \(1024 \times 1024\) (the model’s standard input resolution) would require on the order of \(1024^{4}\) forward passes — clearly infeasible.  
To make the search somehow tractable, we restrict attention to boxes whose center coincides with the object’s center, so we only vary the pair \((h, w)\).  
Even this reduced search amounts to roughly one million inferences — still costly, but manageable for some instances.

\begin{figure}[ht!]
    \centering
    \includegraphics[width=1.0\columnwidth]{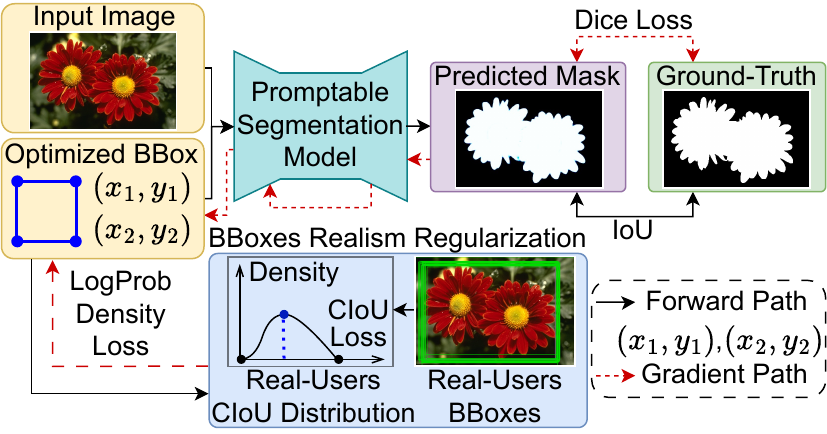}
    \caption{BREPS optimization pipeline.}
    \label{fig:optimization_pipeline}
\end{figure}

For every bounding box, we obtain the model’s prediction and compute the IoU between the prediction and the ground‑truth mask.  
Next, for each pixel we plot the IoU obtained by the bounding box whose corner lies at that pixel. Since there are four such corner points, the resulting visualization (heatmaps for SAM~\cite{kirillov2023segment} in Fig.~\ref{fig:clickbait}, Fig.~\ref{fig:heatmaps}) is symmetric around the object’s center. 

We observed that the IoU heatmaps contain plateaus of high quality that abruptly drop off at certain boundaries. Moreover, even within the high-IoU (red) regions, there are ‘holes’ where the quality is significantly lower. In other words, there exist bounding boxes that differ by only one pixel in their coordinates yet yield significantly different segmentation quality. In an ideal scenario, the model’s output quality would be consistent for all reasonable user-drawn prompts (i.e., the heatmap would be a uniformly red area). This consistency would ensure user satisfaction in practical downstream applications.

\subsection{BREPS Evaluation Protocol}
Exhaustive search over all possible bounding boxes is intractable for large‑scale
datasets. We therefore reformulate the problem as a white‑box adversarial attack in the space of bbox prompts.  
We observed that promptable segmentation models are differentiable with respect
to their input prompts. Thus, we keep the image and the model weights frozen and
perform gradient descent directly in the space of bboxes coordinates. The optimization pipeline is illustrated in Fig.~\ref{fig:optimization_pipeline}. The gradient of the loss function passes through the attacked model and, adjusted with the realism regularizer, shifts the box coordinates accordingly.

We used the same datasets as in Sec.~\ref{datasets}. Since we are no longer limited by the cost of human labeling, we included all instances from GrabCut, Berkeley, DAVIS, COCO-MVal, TETRIS, ACDC, and MedScribble, sampled 1,000 instances from ADE20K and PASCAL-VOC2012, and 503 instances from BUID (benign and malignant).

\paragraph{Realism constraint.}
Naively optimizing the segmentation loss often produces boxes that miss the object entirely, trivially lowering models' performance. To prevent such degenerate solutions, we introduce a realism term based on the Gamma density fitted in Sec.~\ref{real_user_behaviour}.

The overall objective is difference of DICE loss~\cite{dice1945measures} and $\lambda$-scaled log-probability density (Sec.~\ref{real_user_behaviour}, $\star$) of real-users CIoU-Loss distribution.
For the IoU-maximizing attack, we negate the DICE term. We set $\lambda=0.1$, choosing the Pareto‑optimal trade‑off between IoU degradation and log-probability realism (refer to the Supplementary).

\paragraph{Optimization details.}
First, we initialize the optimized box as the \textit{tight bbox} in \((x_1, y_1, x_2, y_2)\) parametrization. Then, we run 50 steps of the Adam optimizer to obtain realistic yet challenging boxes, taking under 2 seconds on an NVIDIA Tesla A100 for the SAM ViT-B model. Additionally, we apply a \textit{clip} operation to ensure the bbox coordinates are inside the image and handle the edges in order (e.g. \(x_1 < x_2\)). Ablation on the number of steps is provided in the Supplementary. Since the optimizer operates in prompt space, whose scale varies between models (owing to different input resolutions), we linearly rescale the learning rate (\(lr=9\)) with respect to the $1024\times1024$ input size of SAM.

\paragraph{Evaluation metrics.}
We adapt click-based metrics from \cite{moskalenko2024tetris} for robustness evaluation on bounding boxes:

\begin{itemize}
  \item \textbf{IoU‑Tight@BBox}: IoU with the \textit{tight bbox} prompt;
  \item \textbf{IoU‑Min/Max@BBox}: IoU on the worst/best‑case bbox found by the
        quality‑decreasing/increasing attack;
  \item \textbf{IoU‑$\Delta$@BBox}: robustness metric defined as the difference between Max and Min attacks.
\end{itemize}

\subsection{Discussion}
Evaluation results on several datasets are shown in Tab.~\ref{tab:breps_attack}. More datasets are provided in the Supplementary. Based on the results, we made several conclusions:
\begin{itemize}
\item Following the results of our real-users study (Fig.~\ref{fig:clickbait}, Fig.~\ref{fig:heatmaps}, Fig.~\ref{fig:user_bboxes}, Tab.~\ref{tab:general_users}, Tab.~\ref{tab:medical_users}) and robustness evaluation (Tab.~\ref{tab:breps_attack}), we can conclude that state-of-the-art interactive segmentation models are \textbf{extremely sensitive to the bounding box prompt fluctuations}.
\item In the real-users study (Fig.~\ref{fig:user_bboxes}, Tab.~\ref{tab:general_users}, Tab.~\ref{tab:medical_users}) we also observed a strong spread in quality between user bboxes; on average, this \textbf{spread is about 15\% of the IoU quality}.
\item During the exhaustive search, we found (Fig.~\ref{fig:heatmaps}) that there exist \textbf{neighboring bbox positions that differ significantly in quality}. We note that the revealed problem is \textbf{relevant for both general and medical domains}.
\item We observed a significant drop in quality when optimizing for minimization. On average, the \textbf{quality drops by 30\% IoU relative to \textit{tight bbox}}, while according to our optimization strategy, bboxes still lie in the distribution of human-probable bboxes. This indicates potential quality drops when using such models on real people and highlights their overfitting to \textit{tight bounding boxes}.
\item Moreover, we observed that \textbf{\textit{tight bboxes} turned out to be suboptimal in terms of maximum possible quality}, meanwhile they \textbf{overestimate real-world performance}. As a result of our optimization for IoU maximization, the quality of the models can be \textbf{increased by around 3\% IoU} on average across all 10 datasets.
\end{itemize}

\section{Conclusion}
In this work, we explore the robustness of promptable segmentation models. Firstly, we gathered 25,000 real bboxes from 2,500 annotators using crowdsourcing. We also provided statistics and described the distribution of the obtained real-users bboxes. On these prompts, we evaluate state-of-the-art models and observe a large quality spread between users. We observed this effect in both general and medical segmentation domains. Then we performed an exhaustive search sweeping millions of plausible bboxes per instance, and revealed IoU gaps even for neighboring pixels. Finally, we proposed a white-box BREPS attack, which maintains the realism of optimized bboxes and efficiently finds adversarial prompts for the minimization and maximization of segmentation quality. We also formulated a robustness score and carried out a large-scale comparison of 15 models on 10 datasets from general to medical segmentation domains.

\section*{Acknowledgements}
This work was supported by the The Ministry of Economic Development of the Russian Federation in accordance with the subsidy agreement (agreement identifier 000000C313925P4H0002; grant No 139-15-2025-012).

\bibliography{main}
\end{document}


\maketitle

\section{Real-User Study}

\subsection{Medical Models on Real-Users}

We tested specialized medical models on real users and only on medical datasets. Results for ScribblePrompt~\cite{wong2024scribbleprompt}, SAM-Med2D~\cite{sam-med2d}, MedSAM~\cite{medsam} provided in Tab.~\ref{tab:medical_users}.

\begin{table}[h]
\centering
\fontsize{9pt}{15pt}\selectfont
\tabcolsep=1.5pt
\begin{tabular}{c|c|ccc}
\toprule
Method                     & Backbone & ACDC        & BUID        & MedScribble \\ \hline
ScribPrompt             & ViT-B    & 30.06±7.77 & 33.49±\phantom{0}8.39 & 19.58±4.67 \\ \hline
SAM-Med2D                  & ViT-B    & 73.89±9.61 & 41.55±11.22 & 39.64±9.41 \\ \hline
MedSAM                     & ViT-B    & 40.25±8.05 & 63.37±\phantom{0}7.83 & 49.19±7.44 \\ \bottomrule
\end{tabular}
\caption{Medical promptable segmentation models' performance on real-user bounding boxes on medical segmentation datasets. The standard deviation was computed across 50 users and averaged over all images in the datasets.}
\label{tab:medical_users}
\end{table}

\subsection{Bounding-Box Quality Analysis}

We collected 2,500 real-users bboxes for each of the 10 datasets. To measure the quality of bboxes relative to \textit{tight bboxes} we chose the CIoU~\cite{zheng2020distance} since in addition to IoU it takes into account the distances between centers and aspect ratios of bboxes, providing smooth gradients even in the case of non-overlapping bboxes.

\begin{figure*}[ht!]
    \centering
    \includegraphics[width=1.0\textwidth]{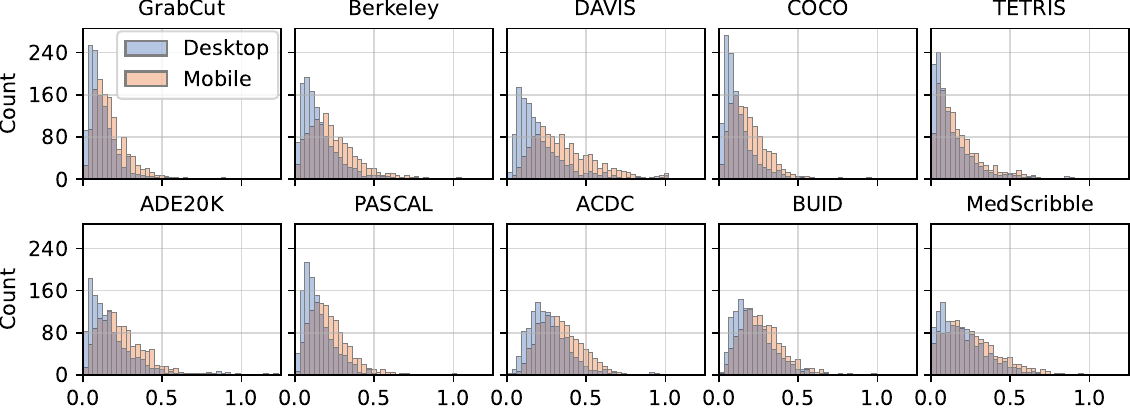}
    \caption{Per-dataset distributions of CIoU-Loss between real-users bboxes and Tight bboxes from ground-truth masks. One can see that the distributions vary between datasets, which tells us that the target instance itself also affects the complexity of the bbox. It is also observed in all datasets that the bboxes on mobile devices are more shifted from the tight ones compared to desktop devices.}
    \label{fig:bboxes_user_distribution_all_datasets}
\end{figure*}

Fig.~\ref{fig:bboxes_user_distribution_all_datasets} shows the distributions between the datasets, divided across mobile and desktop devices. One can see that bboxes on mobile devices on average have lower quality than on desktop in terms of CIoU-Loss. Also, the distribution changes from dataset to dataset. To make our methodology universal and simple, we neglected these differences and combined all datasets and devices and fitted a single Gamma distribution. However, a more precise estimate for each dataset or device is possible in future work.

\begin{figure*}[ht!]
    \centering
    \vspace{0.77cm}
    \includegraphics[width=1.0\textwidth]{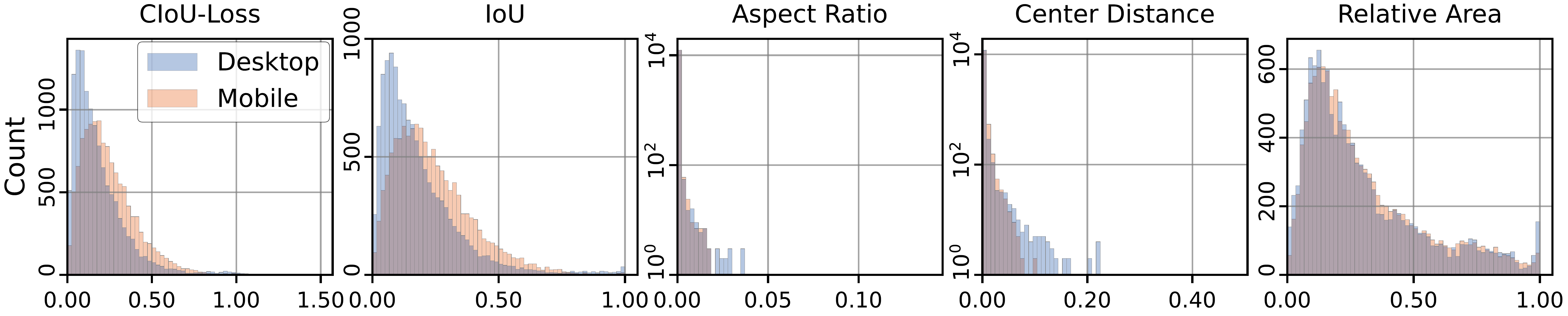}
    \caption{Distributions based on our real-user study, separated between annotators with Desktop and Mobile devices. From left to right: CIoU-Loss; its components --- IoU, Aspect Ratio, Center Distance penalty components; Relative BBox Area normalized by image size. We observed cross-device differences in CIoU and its components. Relative areas remain similar, while more often exceeding the highest value on desktop devices.}
    \label{fig:per_component}
\end{figure*}

Fig.~\ref{fig:per_component} shows the distributions of CIoU-Loss, its components, and relative bbox area distribution across Mobile and Desktop devices. We observe that Mobile bboxes are far from the Tight bboxes than Desktop boxes by CIoU and espesially its IoU component. Aspect ratio and Center Distance penalty have a greater values with Desktop devices. Relative area distributed similarly, while achieving examples with the highest values (bbox covering full image) on Desktop devices.

\begin{figure}[t!]
  \centering
  \includegraphics[width=1.0\columnwidth]{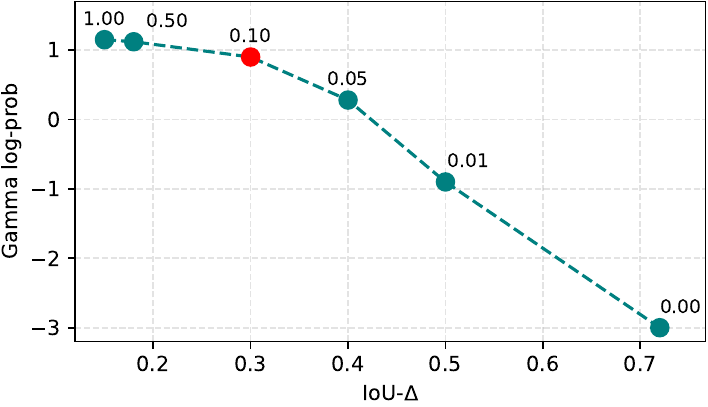}
  \caption{Ablation study on the lambda tradeoff parameter for bboxes. We select 0.1 value as a good trade-off between IoU delta and bboxes realism (Gamma PDF).}
  \label{fig:bbox_pareto}
\end{figure}

\section{Proposed BREPS Attack}

\subsection{Optimizer Learning Rate}

During the BREPS attack we directly optimize the prompt space, which means that the learning rate directly determines how many pixels the box will move per step. However, the models may have different input images and the step of 10 pixels for the 256 and 1024 pixels images differs significantly. To align all the models we used a learning rate = 9 and scaling for the most popular resolution of 1024 pixels. The active learning rate is calculated as follows: $LR=\frac{9 \sqrt{H^2+W^2}}{1024 \sqrt{2}}$, where $H,W$ --- image height and width in pixels respectively.

\subsection{Number of optimization steps}

We investigated the optimal number of steps during optimization with a simple experiment --- we looked at how the average gradient rate changed from step to step and found that at 50 steps the changes no longer exceeded 1\% of the value, so we chose 50 optimization steps as the optimal value in BREPS attack.

\subsection{Regularizer Strength Selection}
\label{sec:regularizer}
We want to optimize two loss functions at the same time --- one is designed to change the segmentation mask at the output of the promptable model, and the other is designed to make the bboxes at the input realistic.

As the first loss function, we choose DICE score~\cite{dice1945measures}, and the second loss is the logarithm of the probability of our proposed real-users CIoU based distribution. The problem arises of balancing these terms in order to obtain simultaneously realistic and at the same time adversarial prompts, which affects models performance.

To choose $\lambda$ tradeoff parameter, we run BREPS with $\lambda$ from 0.0 (no regularization) to 1.0 (hard regularization) and compute the average Gamma log-PDF with average IoU-$\Delta$ over samples from 10 datasets (50 samples from each dataset in the real-users study) for SAM-B~\cite{kirillov2023segment} model. Obtained results provided in Fig.~\ref{fig:bbox_pareto}. Finally, we selected $\lambda=0.1$ as an optimal value since it provides a highly realistic log-PDF value and already provides a significant IoU-$\Delta$.

\subsection{Suitability of Proposed Regularizer}
\label{sec:regularizer_suit}
\begin{figure*}[ht!]
    \centering
    \includegraphics[width=0.54\textwidth]{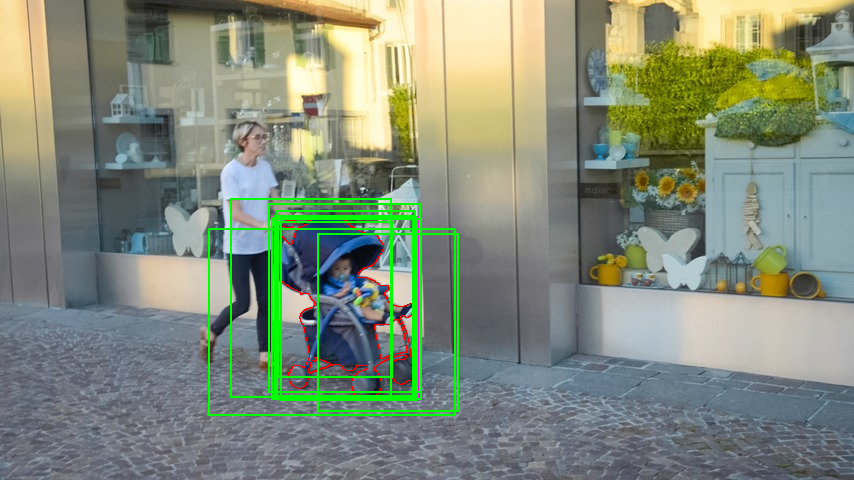}
    \includegraphics[width=0.455\textwidth]{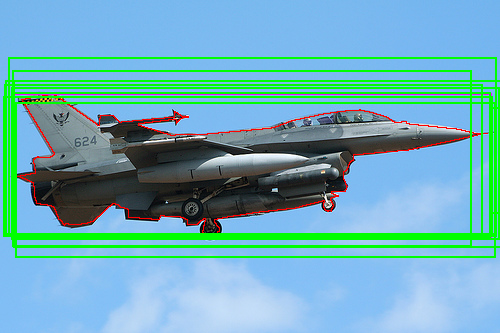}
    \caption{Sampled bboxes are displayed in green, Ground Truth outline is displayed in red.}
    \label{fig:regularizer_sample_vis}
\end{figure*}

To qualitatively assert that the proposed regularizer indeed promotes the realism of adversarial bboxes we sample from the corresponding density using Metropolis Adjusted Langevin Algorithm~\cite{MALA}. Examples provided in Fig.~\ref{fig:regularizer_sample_vis}.

Despite the visual plausibility of samples, such auxiliary density is uniform over bboxes that have equivalent complete-Intersection-over-Union, which may not be the case for an unknown real-user-induced distribution. We perform a multivariate Kolmogorov-Smirnov test~\cite{KSTEST} in order to quantify the level of realism provided by our model distribution. With KS statistic = 0.008 and p-value = 0.3922 we cannot tell Langevin samples apart from real-user bboxes, which justifies the use of our regularizer.

Additionally, we compared our method with~\cite{huang2024robust} bounding boxes generation strategy (with 30\% jitter) and observed IoU-$\Delta=0.443$, log-PDF $=-0.051$, which corresponds to much lower bbox realism than optimization with the proposed regularizer.

\begin{figure*}[ht!]
    \centering
    \includegraphics[width=1.0\textwidth]{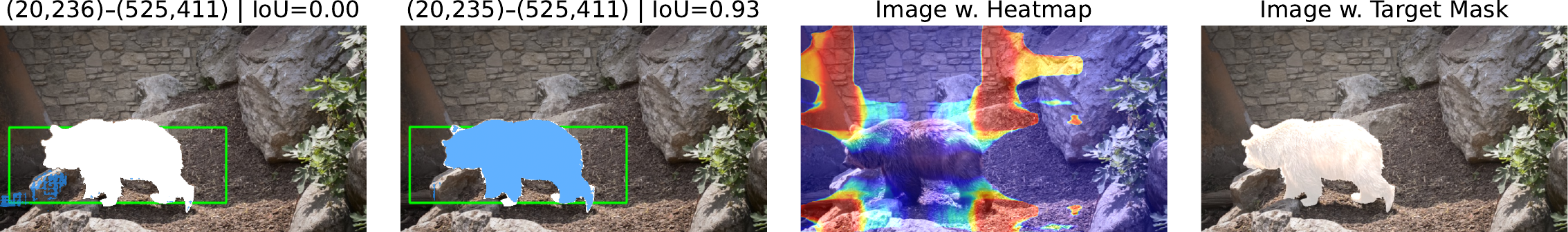}
    \vspace{-0.3cm}
    \\
    \includegraphics[width=1.0\textwidth]
    {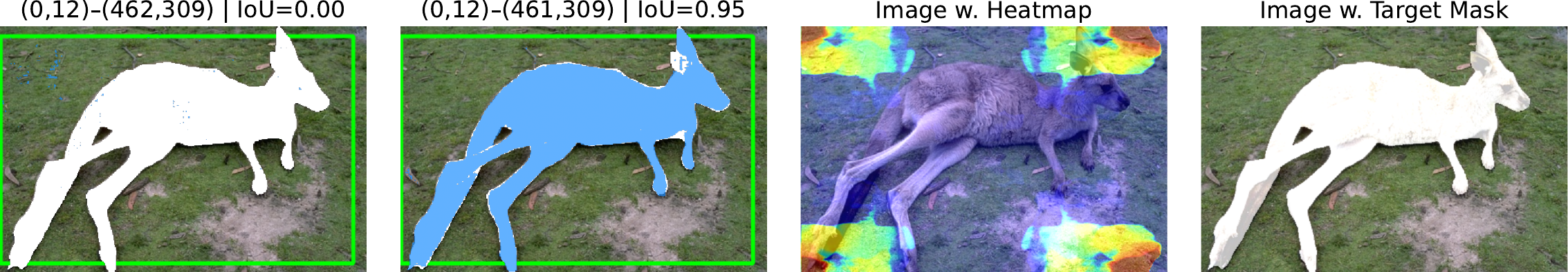}
    \vspace{-0.3cm}
    \\
    \includegraphics[width=1.0\textwidth]
    {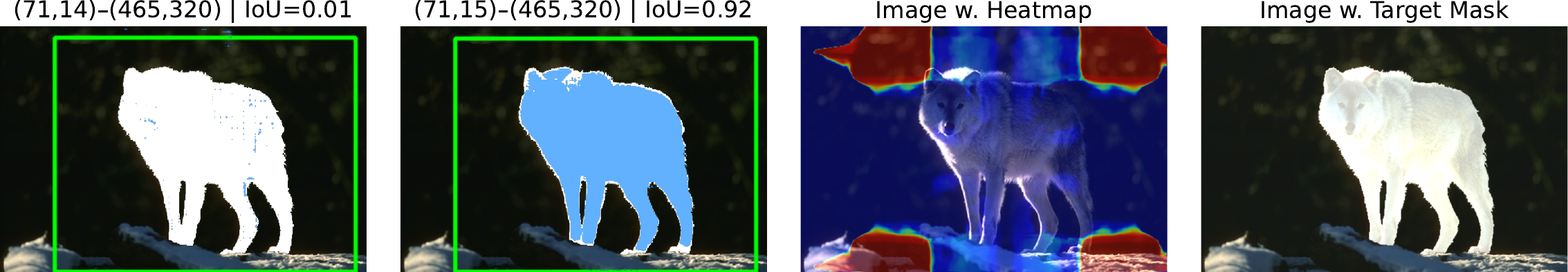}
    \vspace{-0.3cm}
    \\
    \includegraphics[width=1.0\textwidth]
    {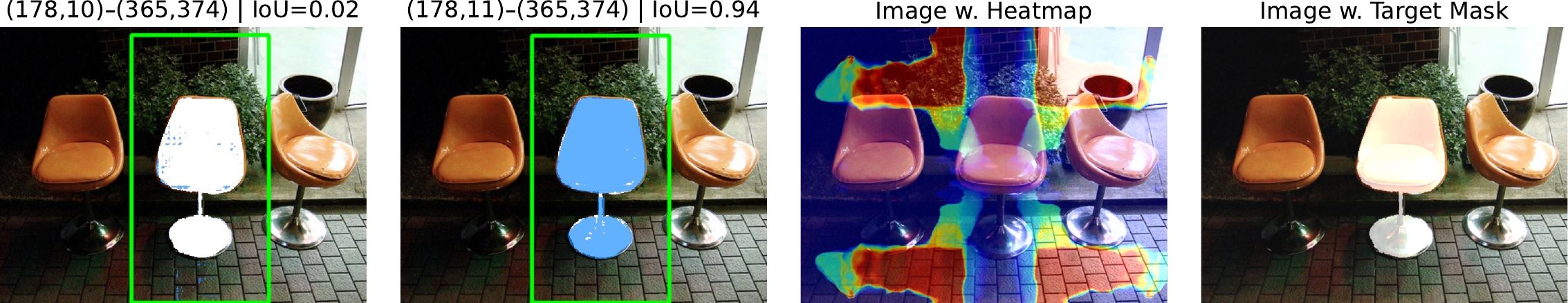}
    \vspace{-0.3cm}
    \\
    \includegraphics[width=1.0\textwidth]
    {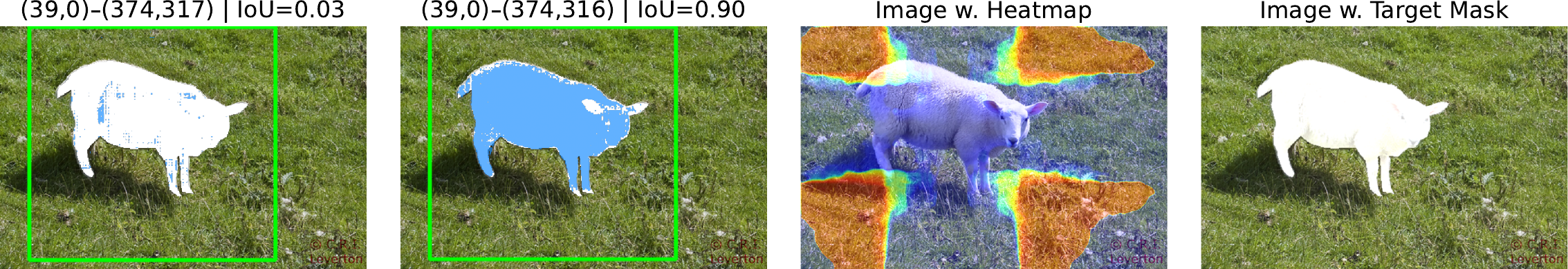}
    \vspace{-0.3cm}
    \\
    \includegraphics[width=1.0\textwidth]
    {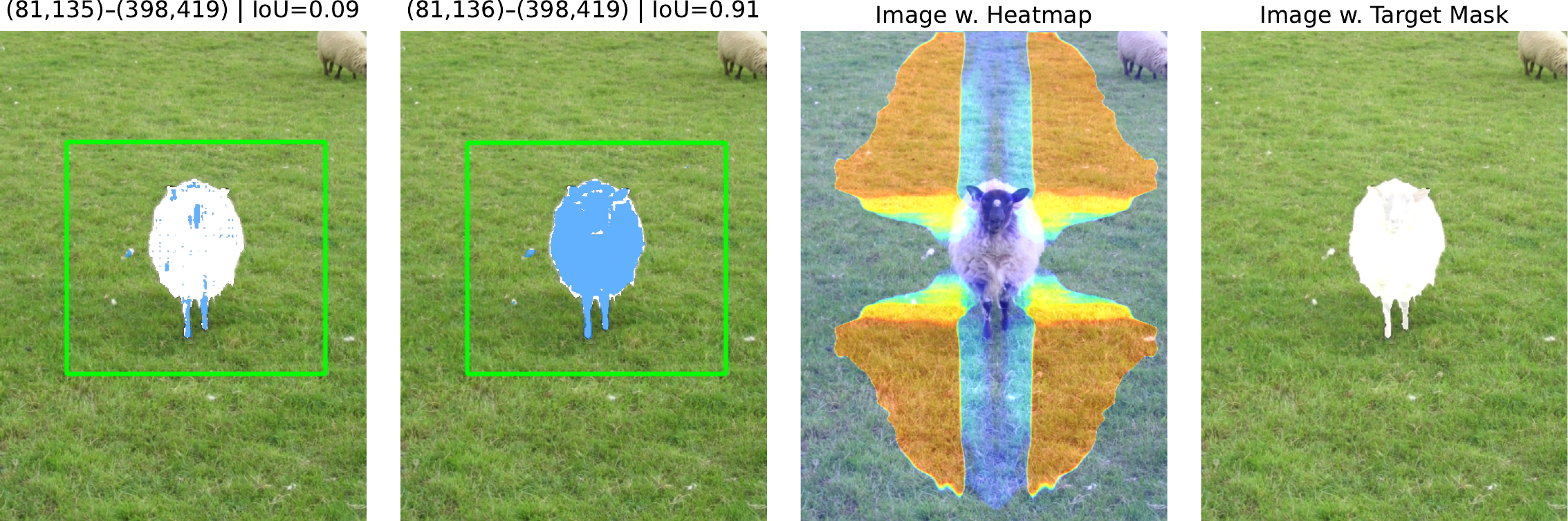}
    \caption{Examples of neighboring bboxes with 1 pixel shifts, which provide significantly different segmentation masks.}
    \label{fig:adj_bboxes1}
\end{figure*}

\begin{figure*}[ht!]
    \centering
    \includegraphics[width=1.0\textwidth]{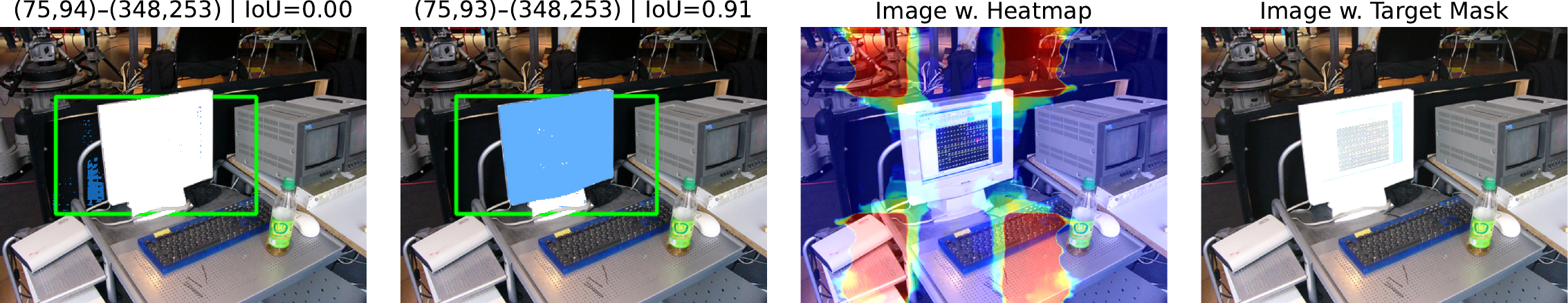}
    \vspace{-0.38cm}
    \\
    \includegraphics[width=1.0\textwidth]
    {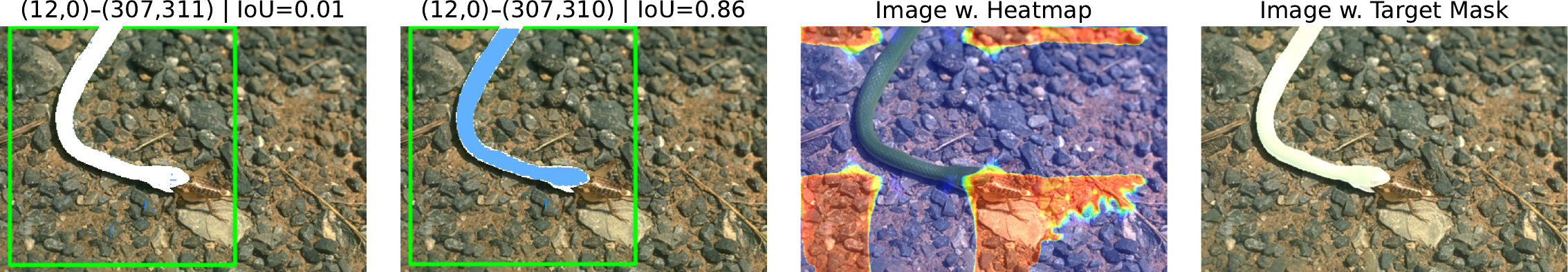}
    \vspace{-0.39cm}
    \\
    \includegraphics[width=1.0\textwidth]
    {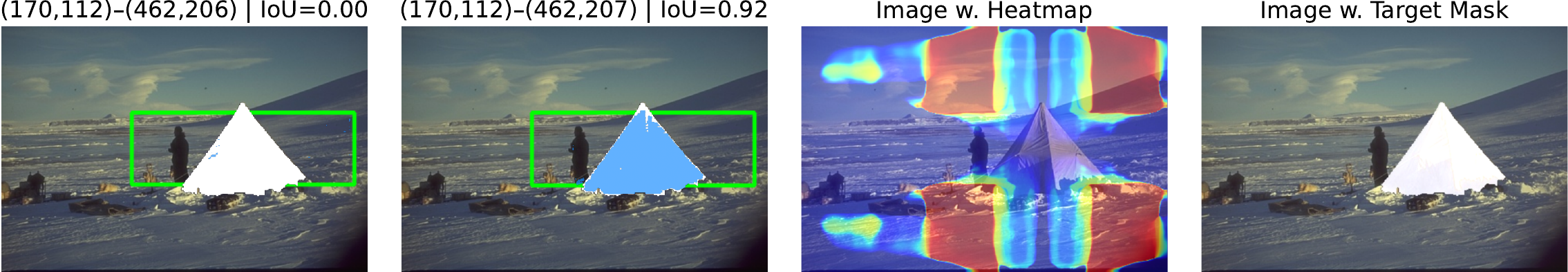}
    \vspace{-0.39cm}
    \\
    \includegraphics[width=1.0\textwidth]
    {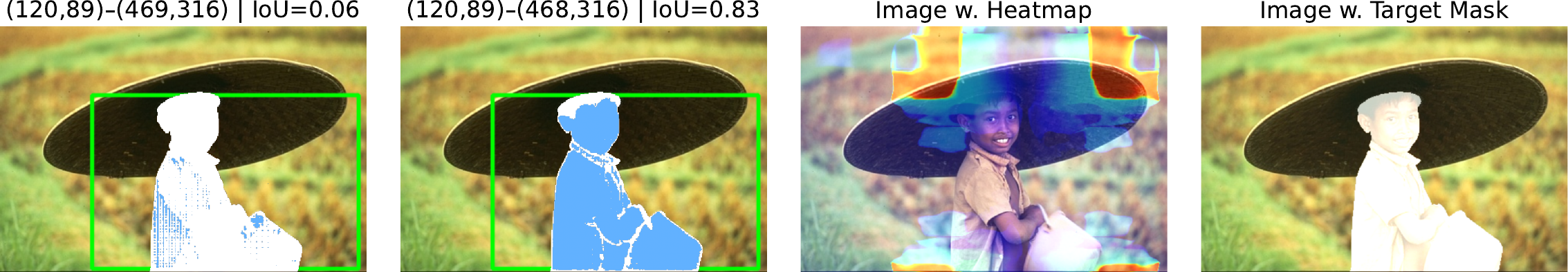}
    \vspace{-0.39cm}
    \\
    \includegraphics[width=1.0\textwidth]
    {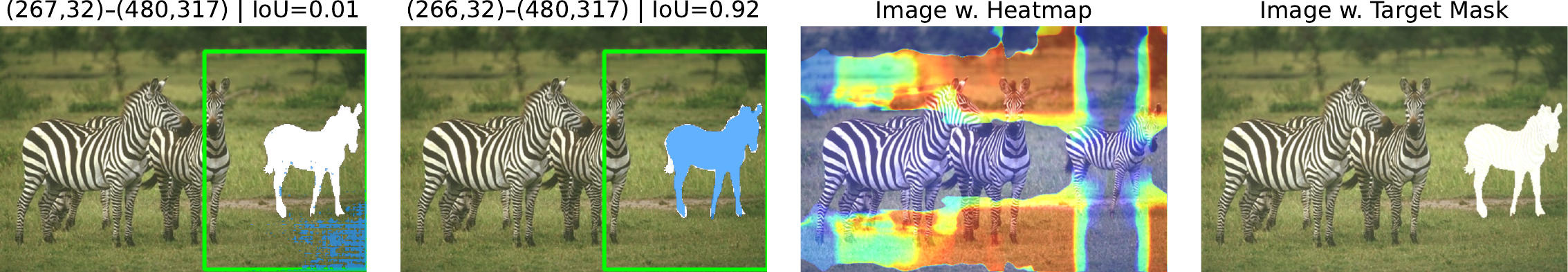}
    \vspace{-0.39cm}
    \\
    \includegraphics[width=1.0\textwidth]
    {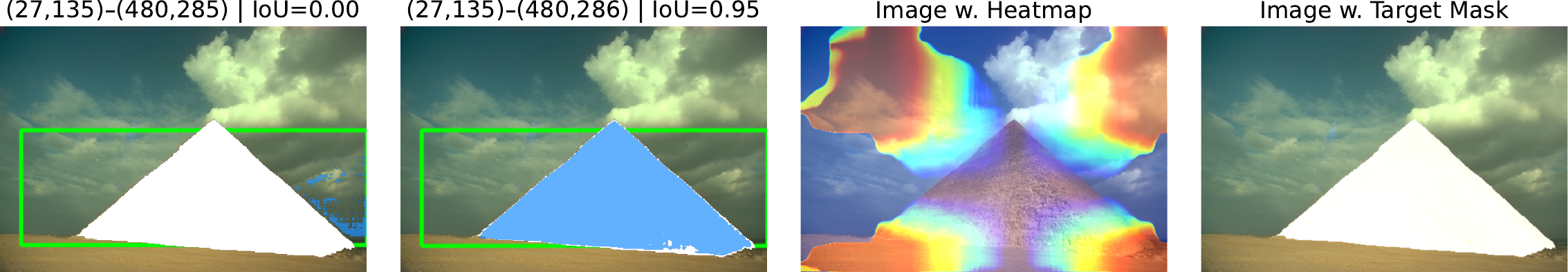}
    \vspace{-0.39cm}
    \\
    \includegraphics[width=1.0\textwidth]
    {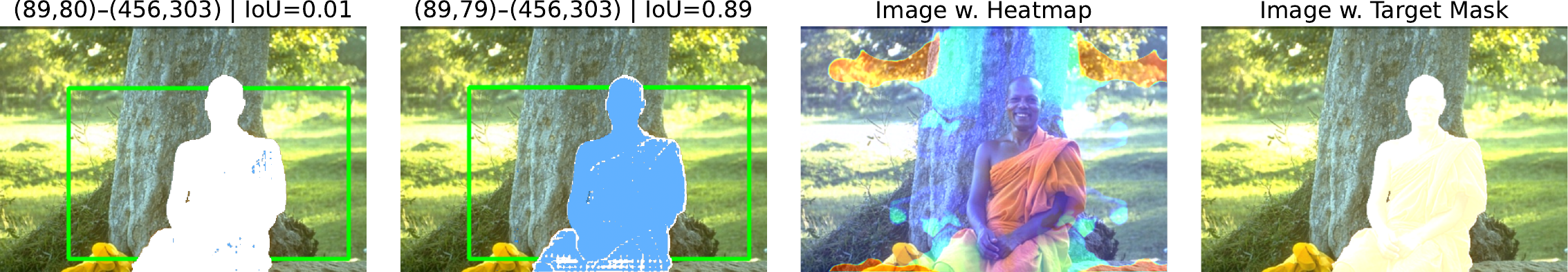}
    \caption{Examples of neighboring bboxes with 1 pixel shifts, which provide significantly different segmentation masks.}
    \label{fig:adj_bboxes2}
\end{figure*}

\begin{figure*}[ht!]
    \centering
    \includegraphics[width=0.975\textwidth]{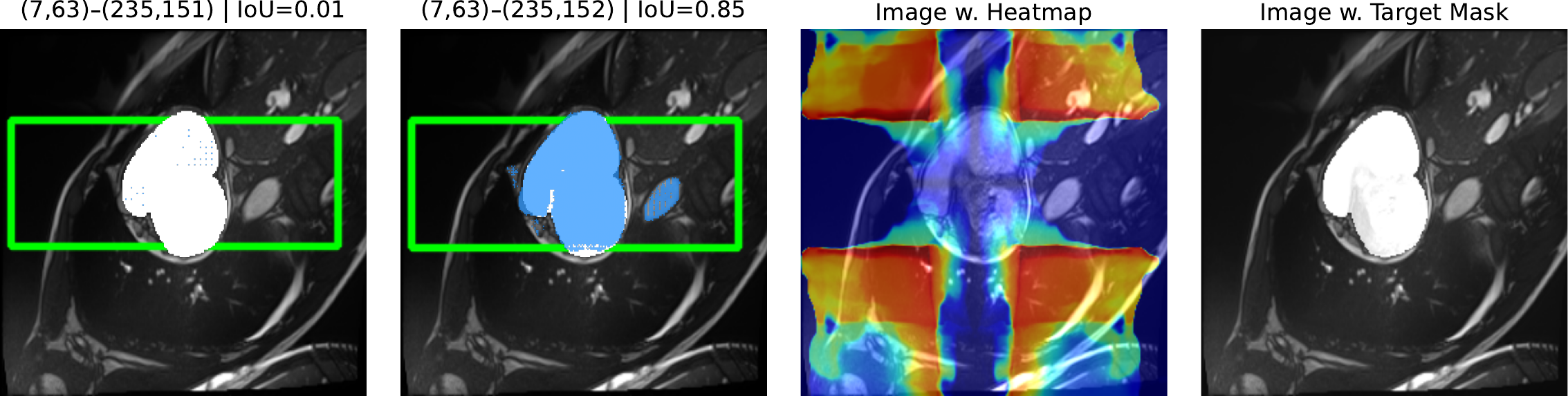}
    \\
    \includegraphics[width=0.975\textwidth]
    {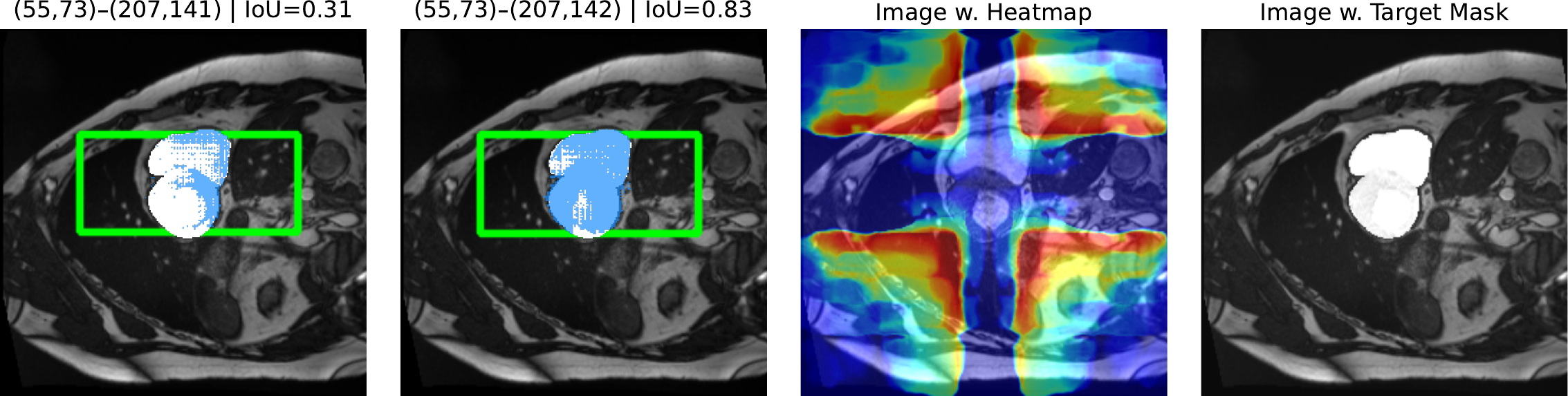}
    \\
    \includegraphics[width=0.975\textwidth]
    {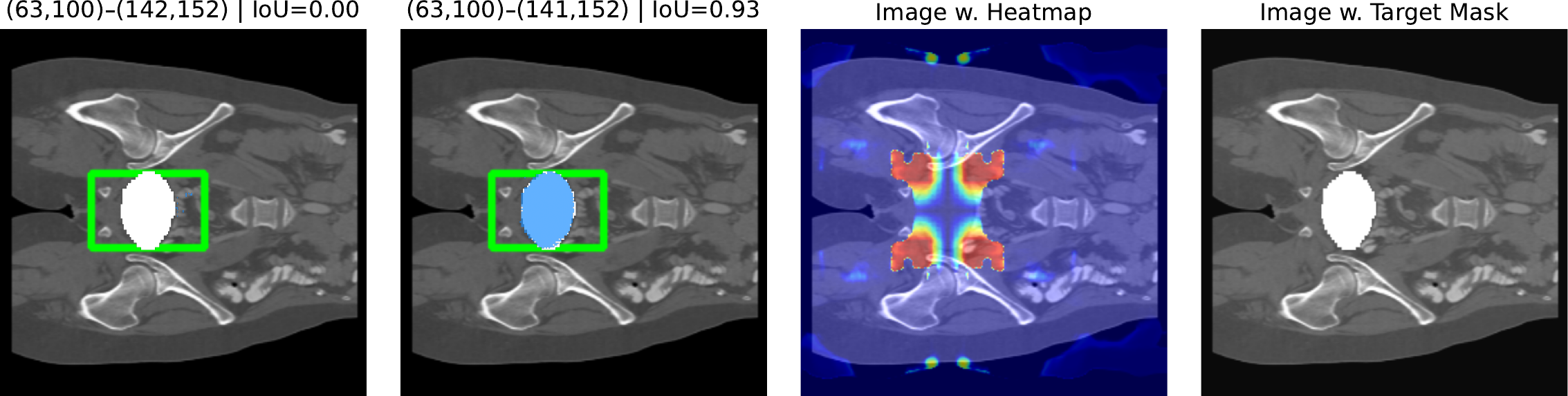}
    \\
    \includegraphics[width=0.975\textwidth]
    {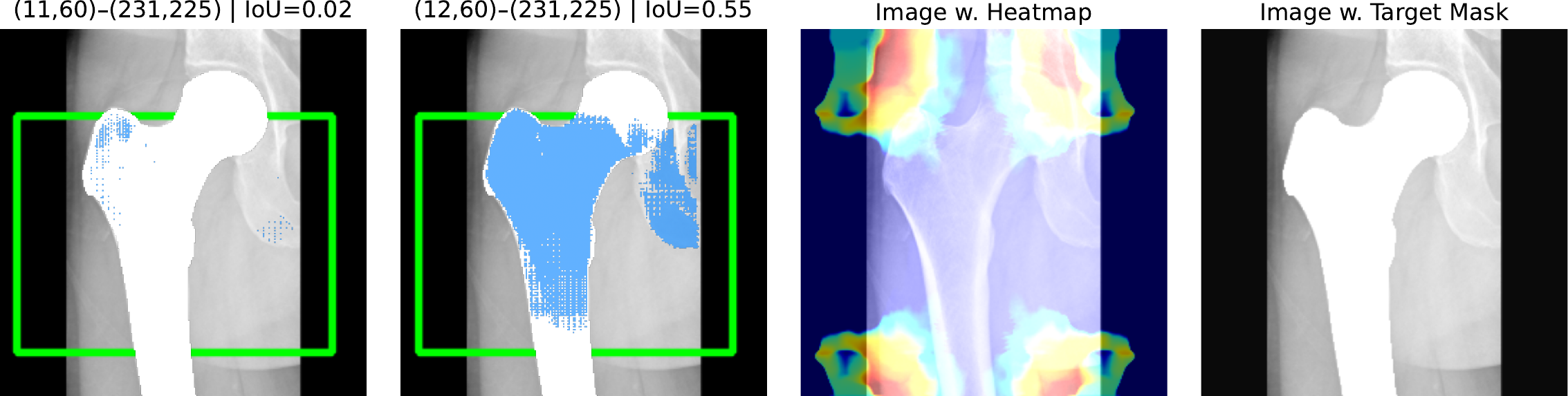}
    \\
    \includegraphics[width=0.975\textwidth]
    {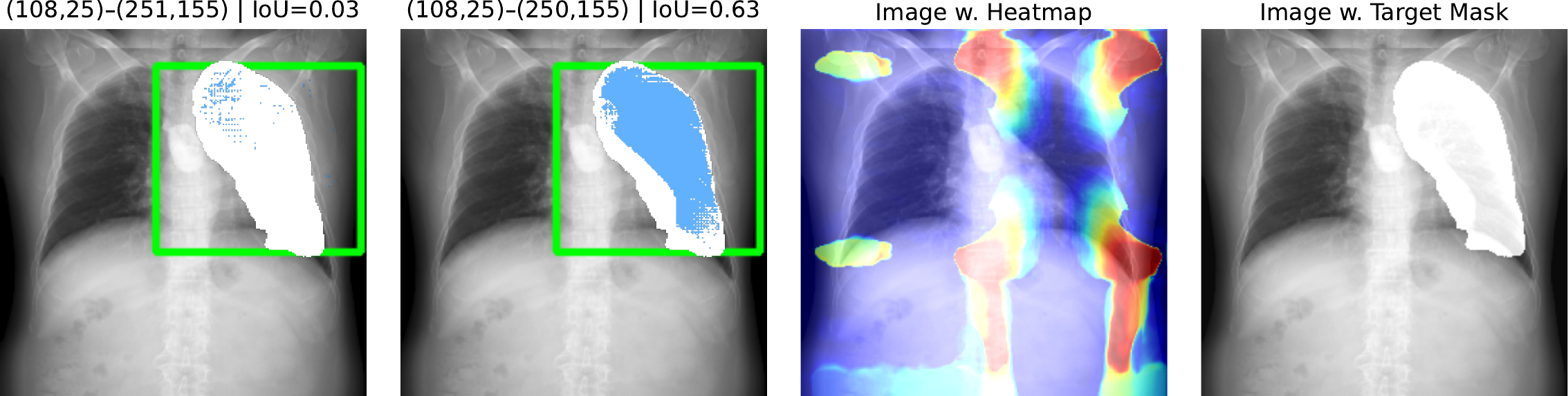}
    \caption{Examples of neighboring bboxes with 1 pixel shifts, which provide significantly different segmentation masks.}
    \label{fig:adj_bboxes3}
\end{figure*}

\subsection{All Datasets Results}

In this section, we provide evaluation results for the remaining datasets and average results across all datasets (for medical models we average only the medical datasets results). Obtained results privided in Tab.~\ref{tab:breps_attack1},~\ref{tab:breps_attack2},~\ref{tab:user_study1},~\ref{tab:user_study2},~\ref{tab:user_study3},~\ref{tab:user_study4}. 

We observed that the method with fewer quality drops and higher maximum and \textit{tight bbox} quality is SAM-HQ and SAM-HQ2 ~\cite{ke2023segment}. From the medical models, we can note that the most performing method is SAM-Med2D~\cite{sam-med2d}, while the most robust is MedSAM~\cite{medsam}.

In Fig.~\ref{fig:adj_bboxes1},~\ref{fig:adj_bboxes2},~\ref{fig:adj_bboxes3}, we provide the examples of adjacent bounding boxes (e.g. different in a single coordinate by a single pixel), which have high variation in segmentation resulting segmentation mask and quality. Such examples can be easily found with our heatmap generation strategy by analysing differences in heighbouring heatmap pixel values.

\begin{table*}[ht!]
\centering
\fontsize{9pt}{14pt}\selectfont
\tabcolsep=2.29pt
\begin{tabular}{c|c|cccccccccccccccccccc}
\toprule
\multirow{2}{*}{Methods}   & \multirow{2}{*}{Backbone} &  & \multicolumn{4}{c}{GrabCut}   &  & \multicolumn{4}{c}{Berkeley}  &  & \multicolumn{4}{c}{DAVIS}     &  & \multicolumn{4}{c}{TETRIS}    \\ \cline{4-7} \cline{9-12} \cline{14-17} \cline{19-22}
                           &                           &  & Tight & Min   & Max   & Delta &  & Tight & Min   & Max   & Delta &  & Tight & Min   & Max   & Delta &  & Tight & Min   & Max   & Delta \\ \cline{1-2} \cline{4-7} \cline{9-12} \cline{14-17} \cline{19-22}
MobileSAM                  & ViT-Tiny                  &  & 94.77 & \textbf{96.21} & 97.19 & \textbf{0.99} &  & 89.85 & 80.89 & 93.60 & 12.72 &  & 85.47 & 73.16 & 88.04 & 14.88 &  & 91.01 & 61.15 & 92.90 & 31.75 \\
\cline{1-2} \cline{4-7} \cline{9-12} \cline{14-17} \cline{19-22}
\multirow{3}{*}{SAM}       & ViT-B                     &  & 91.99 & 70.68 & 95.34 & 24.66 &  & 91.28 & 69.29 & 93.75 & 24.46 &  & 84.93 & 62.67 & 89.03 & 26.36 &  & 91.39 & 52.20 & 93.65 & 41.44 \\
                           & ViT-L                     &  & 94.57 & 85.52 & 96.44 & 10.93 &  & 91.13 & 81.33 & 92.66 & 11.33 &  & 85.17 & 67.34 & 87.33 & 19.99 &  & 92.49 & 62.74 & 94.12 & 31.38 \\
                           & ViT-H                     &  & \textbf{95.24} & 83.20 & 97.08 & 13.88 &  & 91.31 & 82.85 & 93.73 & \underline{10.88} &  & 85.80 & 70.84 & 88.36 & 17.52 &  & 92.50 & 60.79 & 93.86 & 33.06 \\
\cline{1-2} \cline{4-7} \cline{9-12} \cline{14-17} \cline{19-22}
\multirow{3}{*}{SAM-HQ}    & ViT-B                     &  & 94.26 & 82.18 & 97.80 & 15.61 &  & \underline{93.72} & 82.24 & \underline{95.35} & 13.11 &  & \underline{88.91} & \underline{75.28} & 90.86 & 15.58 &  & 92.89 & 63.72 & 94.82 & 31.09 \\
                           & ViT-L                     &  & 94.34 & 90.67 & 97.79 & 7.12  &  & 92.05 & \underline{83.05} & 94.05 & 11.00 &  & 88.13 & 74.96 & 90.47 & 15.50 &  & 93.68 & \underline{71.97} & 95.36 & \underline{23.39} \\
                           & ViT-H                     &  & \underline{95.20} & \underline{94.42} & 97.92 & \underline{3.50}  &  & 92.73 & \textbf{87.53} & 95.18 & \textbf{7.65}  &  & 88.79 & \textbf{76.95} & 90.67 & \underline{13.73} &  & \underline{93.70} & \textbf{74.07} & \underline{95.50} & \textbf{21.43} \\
\cline{1-2} \cline{4-7} \cline{9-12} \cline{14-17} \cline{19-22}
\multirow{4}{*}{SAM 2.1}   & Hiera-T                   &  & 94.66 & 56.80 & 97.41 & 40.61 &  & 92.00 & 66.09 & 94.40 & 28.31 &  & 88.81 & 52.87 & 90.60 & 37.73 &  & 91.69 & 35.63 & 93.80 & 58.17 \\
                           & Hiera-S                   &  & 93.23 & 52.14 & 96.06 & 43.93 &  & 93.19 & 62.94 & 94.71 & 31.77 &  & 86.23 & 50.08 & 89.85 & 39.77 &  & 91.06 & 30.69 & 93.43 & 62.74 \\
                           & Hiera-B+                  &  & 94.86 & 66.10 & \textbf{97.98} & 31.88 &  & 93.31 & 68.65 & 95.24 & 26.59 &  & 88.05 & 59.42 & 90.25 & 30.83 &  & 92.72 & 37.55 & 94.63 & 57.08 \\
                           & Hiera-L                   &  & 93.57 & 61.65 & 96.65 & 34.99 &  & 92.99 & 65.27 & 93.99 & 28.72 &  & 88.33 & 60.12 & \underline{91.31} & 31.19 &  & 92.07 & 42.16 & 94.34 & 52.17 \\
\cline{1-2} \cline{4-7} \cline{9-12} \cline{14-17} \cline{19-22}
SAM-HQ 2                   & Hiera-L                   &  & 94.75 & 83.19 & \underline{97.93} & 14.73 &  & \textbf{94.99} & 73.80 & \textbf{96.52} & 22.72 &  & \textbf{92.00} & 70.87 & \textbf{93.06} & 22.20 &  & \textbf{94.11} & 43.50 & \textbf{95.80} & 52.30 \\
\cline{1-2} \cline{4-7} \cline{9-12} \cline{14-17} \cline{19-22}
\multirow{3}{*}{RobustSAM} & ViT-B                     &  & 84.62 & 45.73 & 90.07 & 44.33 &  & 83.32 & 44.74 & 86.69 & 41.95 &  & 71.18 & 28.72 & 75.28 & 46.55 &  & 80.26 & 26.44 & 84.33 & 57.89 \\
                           & ViT-L                     &  & 81.02 & 10.36 & 83.91 & 73.56 &  & 64.72 & 6.49  & 70.59 & 64.10 &  & 40.49 & 2.39  & 46.69 & 44.30 &  & 62.81 & 3.88  & 68.44 & 64.56 \\
                           & ViT-H                     &  & 54.91 & 38.30 & 61.62 & 23.33 &  & 43.06 & 29.43 & 48.26 & 18.83 &  & 24.09 & 18.47 & 26.63 & \textbf{8.16}  &  & 46.86 & 25.30 & 51.71 & 26.42 \\
\bottomrule
\end{tabular}
\caption{BREPS attack results on state-of-the-art general promptable segmentation models. In addition to the results in the main paper, here we provide the obtained metrics for the remaining 4 general segmentation datasets. Best results are in \textbf{bold}, the second best is \underline{underlined}.}
\label{tab:breps_attack1}
\end{table*}

\begin{table*}[h!]
\centering
\fontsize{9pt}{14pt}\selectfont
\tabcolsep=2.01pt
\begin{tabular}{c|c|cccccccccccccccccccc}
\toprule
\multirow{2}{*}{Methods}   & \multirow{2}{*}{Backbone} &  & \multicolumn{4}{c}{MedScribble}                                      &  & \multicolumn{4}{c}{ACDC}                                              &  & \multicolumn{4}{c}{BUID}                                              &  & \multicolumn{4}{c}{Average (10 or 3 datasets)}                           \\ \cline{4-7} \cline{9-12} \cline{14-17} \cline{19-22}
                           &                           &  & Tight           & Min             & Max             & Delta          &  & Tight           & Min             & Max             & Delta           &  & Tight           & Min             & Max             & Delta           &  & Tight           & Min             & Max             & Delta           \\ \hline
MobileSAM                  & ViT-Tiny                  &  & 57.18           & 44.12           & 61.04           & 17.81          &  & 50.83           & 25.51           & 72.13           & 46.62           &  & 73.14           & 54.27           & 77.95           & 17.81           &  & 79.46           & 60.87           & 84.37           & 23.01           \\ \cline{1-2} \cline{4-7} \cline{9-12} \cline{14-17} \cline{19-22}
\multirow{3}{*}{SAM}       & ViT-B                     &  & 55.99           & 47.74           & 60.82           & 13.08          &  & 28.47           & 14.79           & 52.33           & 37.54           &  & 77.55           & 55.87           & 81.07  & 25.20           &  & 77.46           & 54.38           & 82.84           & 28.46           \\
                           & ViT-L                     &  & 57.18           & 51.82           & 60.15           & \underline{8.33}  &  & 57.66           & 28.11           & 78.25           & 50.14           &  & 75.39           & \underline{58.99} & 78.73           & 19.74           &  & 80.89           & 61.74           & 84.99           & 23.26           \\
                           & ViT-H                     &  & \underline{58.31} & \underline{52.49} & 62.16           & 9.68           &  & 53.07           & 25.28           & 72.64           & 47.36           &  & 75.28           & 58.86           & 79.05           & 20.19           &  & 80.54           & 63.31           & 84.77           & 21.46           \\ \cline{1-2} \cline{4-7} \cline{9-12} \cline{14-17} \cline{19-22}
\multirow{3}{*}{SAM-HQ}    & ViT-B                     &  & 57.16           & 49.15           & 61.08           & 11.93          &  & 5.00            & 0.00            & 40.04           & 40.04           &  & 75.95           & 55.53           & 80.21           & 24.67           &  & 75.84           & 58.46           & 82.34           & 23.87           \\
                           & ViT-L                     &  & 58.03           & 50.29           & 61.17           & 10.88          &  & 74.89           & 31.20           & 81.22 & 50.02           &  & 75.28           & 58.52           & 79.52           & 21.00           &  & 83.18 & \underline{64.59} & 86.44           & 21.85           \\
                           & ViT-H                     &  & 58.37           & \textbf{53.03}  & 62.29           & 9.26           &  & 64.32           & 14.34           & 74.67           & 60.34           &  & 75.12           & 57.49           & 79.36           & 21.87           &  & 82.28           & \textbf{65.52}  & 85.97           & 20.46           \\ \cline{1-2} \cline{4-7} \cline{9-12} \cline{14-17} \cline{19-22}
\multirow{4}{*}{SAM 2.1}   & Hiera-T                   &  & 59.15           & 47.60           & 63.31           & 15.71          &  & 77.32           & 47.12           & 80.65           & 33.54           &  & 77.63           & 43.90           & 82.23           & 38.34           &  & 83.30           & 47.49           & 86.37           & 38.88           \\
                           & Hiera-S                   &  & \textbf{60.03}  & 46.27           & 64.02           & 17.75          &  & 60.82           & 52.55           & 68.86           & 21.60 &  & 76.53           & 45.74           & 80.64           & 34.89           &  & 81.10           & 45.53           & 84.62           & 39.62           \\
                           & Hiera-B+                  &  & 56.63           & 35.64           & 61.31           & 25.68          &  & 62.61           & 36.34           & 75.47           & 39.13           &  & 80.47           & 47.01           & 83.90           & 36.89           &  & 82.38           & 48.50           & 86.29           & 37.79           \\
                           & Hiera-L                   &  & \underline{59.78} & 38.34           & \underline{64.15} & 25.80          &  & 75.95           & 7.46            & 79.58           & 72.12           &  & \underline{82.71} & 51.95           & \underline{86.10} & 34.15           &  & \underline{83.74} & 46.63           & \underline{86.76} & 40.13           \\ \cline{1-2} \cline{4-7} \cline{9-12} \cline{14-17} \cline{19-22}
SAM-HQ 2                   & Hiera-L                   &  & 59.01           & 45.13           & \textbf{65.01}  & 19.88          &  & \textbf{89.46}  & \textbf{66.60}  & \textbf{91.70}  & 25.10           &  & \textbf{88.44}  & \textbf{71.64}  & \textbf{90.34}  & 18.69           &  & \textbf{87.03}  & 61.61           & \textbf{89.82}  & 28.21           \\ \cline{1-2} \cline{4-7} \cline{9-12} \cline{14-17} \cline{19-22}
\multirow{3}{*}{RobustSAM} & ViT-B                     &  & 48.85           & 28.80           & 51.54           & 22.74          &  & \underline{77.66} & 31.52           & \underline{82.57} & 51.05           &  & 67.50           & 25.75           & 71.33           & 45.58           &  & 74.02           & 30.51           & 78.23           & 48.75           \\
                           & ViT-L                     &  & 31.39           & 7.46            & 38.42           & 30.96          &  & 73.17           & 14.08           & 76.69           & 62.61           &  & 29.16           & 2.29            & 34.77           & 32.48           &  & 55.71           & 5.81            & 61.27           & 55.46           \\
                           & ViT-H                     &  & 24.21           & 21.88           & 26.24           & \textbf{4.37}  &  & 18.87           & 17.37           & 20.02           & \textbf{2.66}   &  & 29.35           & 19.71           & 33.23           & \underline{13.52} &  & 36.40           & 24.03           & 40.36           & \underline{16.01} \\ \cline{1-2} \cline{4-7} \cline{9-12} \cline{14-17} \cline{19-22}
ScribblePrompt             & ViT-B                     &  & 37.53           & 20.10           & 44.88           & 24.78          &  & 15.60           & 4.14            & 44.39           & 40.25           &  & 49.11           & 19.00           & 62.80           & 43.80           &  & 34.08           & 14.41           & 50.69           & 36.28           \\ \cline{1-2} \cline{4-7} \cline{9-12} \cline{14-17} \cline{19-22}
SAM-Med2D                  & ViT-B                     &  & 46.65           & 32.43           & 53.35           & 20.93          &  & 75.07           & 17.48           & 78.68           & 61.20           &  & 71.92           & 12.48           & 79.40           & 66.92           &  & 64.55           & 20.80           & 70.48           & 49.68           \\ \cline{1-2} \cline{4-7} \cline{9-12} \cline{14-17} \cline{19-22}
MedSAM                     & ViT-B                     &  & 46.75           & 42.41           & 47.38           & 19.88          &  & 70.95           & \underline{58.09} & 71.57           & \underline{13.48} &  & 66.86           & 56.61           & 68.14           & \textbf{11.54}  &  & 61.52           & 52.37           & 62.36           & \textbf{14.97}  \\ \bottomrule
\end{tabular}
\caption{BREPS attack results on state-of-the-art promptable segmentation models on 3 medical datasets and average results for 3 datasets for medical segmentation models, as well as all 10 datasets average for general segmentation models. Best results are in \textbf{bold}, the second best is \underline{underlined}.}
\label{tab:breps_attack2}
\end{table*}

\begin{table*}[h!]
\centering
\fontsize{9pt}{14pt}\selectfont
\tabcolsep=3pt
\begin{tabular}{c|c|lccclccclccc}
\toprule
\multirow{2}{*}{Model}         & \multirow{2}{*}{Backbone}        &  & \multicolumn{3}{c}{ACDC}          & \multicolumn{1}{c}{} & \multicolumn{3}{c}{ADE20K}        &  & \multicolumn{3}{c}{BUID}          \\ \cline{4-6} \cline{8-10} \cline{12-14}
                               &                                  &  & Tight & Mobile      & Desktop          &                      & Tight & Mobile      & Desktop          &  & Tight & Mobile      & Desktop          \\ \cline{1-2} \cline{4-6} \cline{8-10} \cline{12-14}
MobileSAM                      & ViT-T  &  & 86.28 & 78.37±11.87 & 80.20±13.33 &                      & \underline{77.15} & \underline{70.51}±23.42 & \textbf{71.89}±23.93 &  & 61.62 & 56.65±17.02 & 56.63±18.87 \\ \cline{1-2} \cline{4-6} \cline{8-10} \cline{12-14}
\multirow{3}{*}{RobustSAM}     & ViT-B  &  & 74.46 & 60.42±16.12 & 63.99±17.15 &                      & 59.92 & 53.49±21.99 & 55.09±23.07 &  & 38.04 & 34.71±18.67 & 35.03±19.43 \\
                               & ViT-L  &  & 45.55 & 42.03±14.32 & 44.15±14.37 &                      & 38.76 & 31.58±21.78 & 34.06±22.45 &  & 16.69 & 15.28±11.21 & 15.50±11.43 \\
                               & ViT-H  &  & 54.84 & 45.75±8.99  & 47.87±9.58  &                      & 56.13 & 52.13±15.68 & 52.67±16.82 &  & 50.82 & 47.24±16.85 & 47.27±17.00 \\ \cline{1-2} \cline{4-6} \cline{8-10} \cline{12-14}
\multirow{3}{*}{SAM}           & ViT-B  &  & 89.75 & 82.74±11.82 & 84.12±13.66 &                      & 68.79 & 61.30±26.98 & 63.04±27.29 &  & 67.39 & 60.86±16.78 & 61.51±17.76 \\
                               & ViT-L  &  & \underline{91.29} & \textbf{83.61}±11.20 & \textbf{84.90}±13.24 &                      & 70.99 & 63.62±25.28 & 65.21±25.92 &  & 67.37 & 63.32±13.68 & 63.37±14.95 \\
                               & ViT-H  &  & 90.44 & 82.42±11.17 & 83.77±12.91 &                      & 68.10 & 63.32±25.17 & 64.04±25.94 &  & 68.95 & 62.74±13.55 & 63.50±15.01 \\ \cline{1-2} \cline{4-6} \cline{8-10} \cline{12-14}
\multirow{4}{*}{SAM 2.1}       & Hiera-T  &  & 90.82 & 78.94±12.58 & 81.53±14.17 &                      & 73.94 & 68.09±24.15 & 69.20±24.89 &  & 68.17 & 60.19±19.75 & 61.46±20.30 \\
                               & Hiera-S  &  & 90.23 & 77.56±11.90 & 80.20±14.44 &                      & 72.50 & 65.79±26.68 & 65.97±27.07 &  & 68.97 & 63.04±17.08 & \underline{68.58}±13.87 \\
                               & Hiera-B+ &  & 91.26 & 79.15±12.92 & 80.19±12.82 &                      & 74.12 & 67.59±24.08 & 67.09±27.64 &  & 71.04 & 61.08±18.30 & 63.36±18.54 \\
                               & Hiera-L  &  & 89.25 & 78.02±12.36 & 81.47±13.36 &                      & 71.51 & 64.75±26.83 & 69.03±25.96 &  & \underline{74.41} & \underline{67.64}±11.70 & 60.11±22.16 \\ \cline{1-2} \cline{4-6} \cline{8-10} \cline{12-14}
SAM 2.1 HQ                     & Hiera-L  &  & 90.72 & 78.28±11.72 & 80.81±13.22 &                      & 76.68 & 68.54±25.93 & 70.80±26.18 &  & \textbf{77.91} & \textbf{68.56}±11.54 & \textbf{69.94}±13.48 \\ \cline{1-2} \cline{4-6} \cline{8-10} \cline{12-14}
\multirow{3}{*}{SAM-HQ}        & ViT-B  &  & 90.85 & 82.52±9.96  & 83.64±12.35 &                      & 74.41 & 67.47±27.46 & 69.26±27.92 &  & 64.36 & 57.91±22.39 & 58.87±22.60 \\
                               & ViT-L  &  & \textbf{91.32} & \underline{82.86}±10.56 & \underline{84.24}±12.62 &                      & \textbf{77.49} & \textbf{70.86}±23.78 & \underline{71.62}±24.51 &  & 68.17 & 63.41±17.21 & 63.38±18.28 \\
                               & ViT-H  &  & 90.22 & 82.26±10.38 & 83.40±12.51 &                      & 72.95 & 68.59±24.59 & 68.76±25.38 &  & 68.89 & 62.93±16.80 & 63.11±18.11 \\ \bottomrule
\end{tabular}
\caption{State-of-the-art promptable segmentation model's  performance variation under diverse human-collected bounding boxes for 3 out of 10 selected general and medical segmentation datasets. Standard deviation values computed across different users and averaged across dataset sample. Additionally we provide IoU, obtained with \textit{tight-bbox} prompt. Best results are in \textbf{bold}, second best are \underline{underlined}.\\ \\ \\}
\label{tab:user_study1}
\end{table*}

\begin{table*}[h!]
\centering
\fontsize{9pt}{14pt}\selectfont
\tabcolsep=3pt
\begin{tabular}{c|c|lccclccclccc}
\toprule
\multirow{2}{*}{Model}         & \multirow{2}{*}{Backbone}        &  & \multicolumn{3}{c}{Berkeley}      & \multicolumn{1}{c}{} & \multicolumn{3}{c}{COCO}          &  & \multicolumn{3}{c}{DAVIS}         \\ \cline{4-6} \cline{8-10} \cline{12-14} 
                               &                                  &  & Tight & Mobile      & Desktop          &                      & Tight & Mobile      & Desktop          &  & Tight & Mobile      & Desktop          \\ \cline{1-2} \cline{4-6} \cline{8-10} \cline{12-14} 
MobileSAM                      & ViT-T  &  & 86.67 & 83.28±14.90 & 84.06±15.45 &                      & 87.93 & 84.18±14.15 & 84.74±15.75 &  & 80.75 & 75.69±21.83 & 77.38±20.75 \\ \cline{1-2} \cline{4-6} \cline{8-10} \cline{12-14} 
\multirow{3}{*}{RobustSAM}     & ViT-B  &  & 79.96 & 73.58±16.33 & 75.73±16.37 &                      & 79.76 & 75.02±15.46 & 76.47±16.95 &  & 65.66 & 58.11±25.14 & 62.60±24.18 \\
                               & ViT-L  &  & 50.48 & 43.48±19.81 & 46.83±18.52 &                      & 58.12 & 51.43±20.75 & 53.96±20.63 &  & 31.31 & 26.00±19.94 & 29.93±19.98 \\
                               & ViT-H  &  & 53.32 & 48.70±20.61 & 49.89±20.98 &                      & 64.71 & 61.21±16.55 & 61.87±17.26 &  & 29.11 & 26.51±16.32 & 27.64±17.29 \\ \cline{1-2} \cline{4-6} \cline{8-10} \cline{12-14} 
\multirow{3}{*}{SAM}           & ViT-B  &  & 86.50 & 80.53±23.35 & 82.12±22.46 &                      & 86.90 & 79.65±24.15 & 82.02±23.11 &  & 84.41 & 78.89±23.13 & 81.57±21.20 \\
                               & ViT-L  &  & 88.73 & 83.99±21.35 & 85.02±21.23 &                      & 88.80 & 84.44±20.30 & 85.88±19.62 &  & 87.41 & 82.92±20.46 & 84.77±19.37 \\
                               & ViT-H  &  & 87.50 & 83.54±22.79 & 84.20±22.37 &                      & 88.62 & 84.59±21.23 & 85.70±20.81 &  & 87.15 & 82.98±21.05 & 84.85±19.22 \\ \cline{1-2} \cline{4-6} \cline{8-10} \cline{12-14} 
\multirow{4}{*}{SAM 2.1} & Hiera-T  &  & 88.95 & 84.81±19.74 & 85.75±21.16 &                      & 88.14 & 83.68±19.37 & 85.55±20.80 &  & 88.50 & 82.38±20.71 & 85.70±20.37 \\
                               & Hiera-S  &  & 90.94 & 85.78±20.35 & 86.49±20.61 &                      & 88.65 & 83.79±20.63 & 85.89±20.88 &  & 88.42 & 80.53±24.62 & \underline{86.73}±19.16 \\
                               & Hiera-B+ &  & 90.09 & 84.06±21.94 & 87.25±19.70 &                      & 89.58 & 84.05±21.50 & 85.41±20.76 &  & 89.33 & 81.09±25.00 & 84.61±20.22 \\
                               & Hiera-L  &  & 90.44 & 85.15±21.06 & 85.80±19.56 &                      & 89.23 & 85.01±20.37 & 85.03±20.06 &  & \underline{90.14} & 84.61±20.83 & 84.93±19.52 \\ \cline{1-2} \cline{4-6} \cline{8-10} \cline{12-14} 
SAM 2.1 HQ               & Hiera-L  &  & \textbf{92.15} & 87.62±17.93 & \textbf{88.77}±17.87 &                      & \textbf{90.88} & 86.04±18.17 & 87.14±19.01 &  & \textbf{90.75} & 84.84±20.62 & \textbf{87.65}±18.39 \\ \cline{1-2} \cline{4-6} \cline{8-10} \cline{12-14} 
\multirow{3}{*}{SAM-HQ}        & ViT-B  &  & \underline{91.32} & \underline{88.21}±13.67 & 88.37±14.45 &                      & 89.94 & 86.33±14.62 & 86.80±16.41 &  & 87.31 & 83.28±18.52 & 84.55±18.76 \\
                               & ViT-L  &  & 90.81 & 87.91±15.86 & 88.27±16.50 &                      & \underline{90.72} & \textbf{87.05}±15.46 & \textbf{87.61}±16.71 &  & 89.04 & \underline{85.66}±17.35 & 86.57±18.29 \\
                               & ViT-H  &  & 90.79 & \textbf{88.45}±16.19 & \underline{88.44}±17.02 &                      & 89.29 & \underline{86.65}±17.32 & \underline{87.15}±18.22 &  & 88.80 & \textbf{85.94}±17.85 & 86.63±18.22 \\ \bottomrule 
\end{tabular}
\caption{State-of-the-art promptable segmentation model's  performance variation under diverse human-collected bounding boxes for 3 out of 10 selected general and medical segmentation datasets. Results provided separately for Mobile and Desktop device types. Standard deviation values were computed across different users and averaged across the dataset sample. Additionally, we provide IoU, obtained with \textit{tight-bbox} prompt. Best results are in \textbf{bold}, second best are \underline{underlined}.}
\label{tab:user_study2}
\end{table*}

\begin{table*}[h!]
\centering
\fontsize{9pt}{14pt}\selectfont
\tabcolsep=3.12pt
\begin{tabular}{c|c|lccclccclccc}
\toprule
\multirow{2}{*}{Model}     & \multirow{2}{*}{Backbone} &  & \multicolumn{3}{c}{Grabcut}                                & \multicolumn{1}{c}{} & \multicolumn{3}{c}{MedScribble}                            &  & \multicolumn{3}{c}{PASCAL}                                 \\ \cline{4-6} \cline{8-10} \cline{12-14} 
                           &                           &  & Tight                          & Mobile      & Desktop          &                      & Tight                          & Mobile      & Desktop          &  & Tight                          & Mobile      & Desktop          \\ \cline{1-2} \cline{4-6} \cline{8-10} \cline{12-14} 
MobileSAM                  & ViT-T                     &  & 93.20                          & 88.58±17.30 & 89.26±17.47 &                      & 51.79                          & 46.03±23.05 & 47.73±24.43 &  & 85.72                          & 81.63±19.33 & 82.47±19.67 \\ \cline{1-2} \cline{4-6} \cline{8-10} \cline{12-14} 
\multirow{3}{*}{RobustSAM} & ViT-B                     &  & 82.52                          & 77.87±15.50 & 78.92±16.15 &                      & 49.50                          & 45.40±20.16 & 46.63±20.89 &  & 77.22                          & 72.43±17.60 & 73.76±18.12 \\
                           & ViT-L                     &  & 70.30                          & 63.41±19.99 & 66.09±18.33 &                      & 28.89                          & 27.34±16.49 & 27.93±16.43 &  & 60.31                          & 54.83±22.67 & 57.33±22.35 \\
                           & ViT-H                     &  & 68.51                          & 64.50±17.07 & 65.12±17.68 &                      & 40.32                          & 36.62±14.08 & 36.92±14.51 &  & 62.61                          & 58.34±16.15 & 59.45±16.91 \\ \cline{1-2} \cline{4-6} \cline{8-10} \cline{12-14} 
\multirow{3}{*}{SAM}       & ViT-B                     &  & 91.95                          & 85.27±23.40 & 87.14±22.46 &                      & 56.77                          & 47.83±27.70 & 49.81±28.63 &  & 86.07                          & 81.71±23.11 & 83.11±22.22 \\
                           & ViT-L                     &  & 93.08                          & 88.24±19.86 & 88.94±20.36 &                      & 59.05                          & 53.82±25.77 & 54.62±26.15 &  & 88.71                          & 86.15±18.95 & 86.15±20.37 \\
                           & ViT-H                     &  & 94.34                          & 88.83±20.08 & 89.39±20.31 &                      & 58.65                          & \underline{53.22}±25.06 & 54.25±25.81 &  & 88.35                          & 86.42±18.76 & 86.03±20.64 \\ \cline{1-2} \cline{4-6} \cline{8-10} \cline{12-14} 
\multirow{4}{*}{SAM 2.1}   & Hiera-T                     &  & 94.45                          & 89.44±18.56 & 90.07±20.51 &                      & 60.52                          & \textbf{53.90}±23.75 & 52.94±24.84 &  & 90.37                          & 86.71±15.08 & 87.36±17.50 \\
                           & Hiera-S                     &  & 94.52                          & 89.72±18.95 & 90.43±19.72 &                      & \textbf{61.31}                 & 53.12±24.42 & \underline{55.16}±26.45 &  & 90.21                          & 85.06±18.51 & 87.47±17.88 \\
                           & Hiera-B+                    &  & 94.85                          & 89.37±19.37 & 91.00±18.26 &                      & 58.92                          & 50.77±23.50 & 55.08±25.49 &  & 90.95                          & 87.01±16.15 & 86.58±17.48 \\
                           & Hiera-L                     &  & 94.00                          & 90.19±18.19 & 90.74±18.60 &                      & \underline{60.86} & 53.21±25.09 & \textbf{55.48}±24.94 &  & \underline{91.08} & 87.06±17.11 & 86.87±17.20 \\ \cline{1-2} \cline{4-6} \cline{8-10} \cline{12-14} 
SAM 2.1 HQ                 & Hiera-L                     &  & \textbf{95.36}                 & \underline{92.31}±15.44 & \textbf{92.40}±16.97 &                      & 60.47                          & 51.95±24.00 & 54.31±25.54 &  & \textbf{91.17}                 & 87.38±15.71 & \underline{87.56}±17.24 \\ \cline{1-2} \cline{4-6} \cline{8-10} \cline{12-14} 
\multirow{3}{*}{SAM-HQ}    & ViT-B                     &  & 94.85                          & 91.56±14.33 & 91.93±15.34 &                      & 57.60                          & 48.64±24.64 & 51.07±24.90 &  & 88.46                          & 86.02±16.14 & 85.96±18.92 \\
                           & ViT-L                     &  & 94.89                          & 92.08±14.43 & \underline{92.21}±15.09 &                      & 57.28                          & 52.54±24.51 & 53.66±25.27 &  & 90.32                          & \textbf{88.31}±15.82 & \textbf{87.63}±18.76 \\
                           & ViT-H                     &  & \underline{94.99} & \textbf{92.44}±14.48 & 92.17±15.66 &                      & 57.85                          & 52.77±23.93 & 53.93±25.09 &  & 90.35                          & \underline{88.09}±15.51 & 87.53±18.11 \\ \bottomrule

\end{tabular}
\caption{State-of-the-art promptable segmentation model's  performance variation under diverse human-collected bounding boxes for 3 out of 10 selected general and medical segmentation datasets. Results provided separately for Mobile and Desktop device types. Standard deviation values computed across different users and averaged across the dataset sample. Additionally, we provide IoU, obtained with \textit{tight-bbox} prompt. Best results are in \textbf{bold}, second best are \underline{underlined}.\\ \\}
\label{tab:user_study3}
\end{table*}

\begin{table*}[h!]
\centering
\fontsize{9pt}{14pt}\selectfont
\tabcolsep=4pt
\begin{tabular}{c|c|lccclccc}
\toprule
\multirow{2}{*}{Model}     & \multirow{2}{*}{Backbone} &  & \multicolumn{3}{c}{TETRIS}                                 & \multicolumn{1}{c}{} & \multicolumn{3}{c}{Average}                        \\ \cline{4-6} \cline{8-10} 
                           &                           &  & Tight                          & Mobile      & Desktop          &                      & Tight          & Mobile      & Desktop                  \\ \cline{1-2} \cline{4-6} \cline{8-10} 
MobileSAM                  & ViT-T                     &  & 76.85                          & 75.80±23.84 & 75.50±25.30 &                      & 84.04          & 79.95±6.20  & 80.76±6.05          \\ \cline{1-2} \cline{4-6} \cline{8-10} 
\multirow{3}{*}{RobustSAM} & ViT-B                     &  & 64.82                          & 61.73±19.52 & 62.39±20.82 &                      & 72.84          & 67.46±9.51  & 69.28±9.12          \\
                           & ViT-L                     &  & 46.19                          & 41.29±21.29 & 42.84±21.59 &                      & 50.78          & 44.58±13.12 & 47.29±12.88         \\
                           & ViT-H                     &  & 55.25                          & 53.06±23.46 & 53.21±24.14 &                      & 55.66          & 52.07±12.54 & 52.84±12.39         \\ \cline{1-2} \cline{4-6} \cline{8-10} 
\multirow{3}{*}{SAM}       & ViT-B                     &  & 75.76                          & 70.49±27.80 & 71.48±27.89 &                      & 82.91          & 76.83±8.19  & 78.64±8.35          \\
                           & ViT-L                     &  & 82.66                          & 79.40±23.64 & 79.75±24.41 &                      & 85.77          & 81.25±8.24  & 82.24±8.00          \\
                           & ViT-H                     &  & 83.90                          & 79.84±24.61 & 79.07±26.29 &                      & 85.42          & 81.36±8.44  & 81.89±8.45          \\ \cline{1-2} \cline{4-6} \cline{8-10} 
\multirow{4}{*}{SAM 2.1}   & Hiera-T                     &  & 80.48                          & 75.32±27.91 & 78.88±28.37 &                      & 86.41          & 81.49±7.36  & 82.69±7.40          \\
                           & Hiera-S                     &  & 82.53                          & 76.40±28.14 & 78.74±28.66 &                      & 86.83 & 81.01±7.92  & 82.83±7.98 \\
                           & Hiera-B+                    &  & 85.34                          & 77.89±27.57 & 77.88±27.86 &                      & 87.76          & 81.58±7.21  & 83.22±7.04          \\
                           & Hiera-L                     &  & 82.70                          & 78.74±27.17 & 76.43±28.93 &                      & 87.02          & 82.21±8.43  & 83.10±8.34          \\ \cline{1-2} \cline{4-6} \cline{8-10} 
SAM 2.1 HQ                 & Hiera-L                     &  & \textbf{86.99}                 & \underline{81.50}±24.96 & 81.91±26.15 &                      & \textbf{89.14} & 84.03±7.57  & \underline{85.18}±7.05          \\ \cline{1-2} \cline{4-6} \cline{8-10} 
\multirow{3}{*}{SAM-HQ}    & ViT-B                     &  & 83.65                          & 80.96±18.42 & 80.65±20.28 &                      & 87.14          & 83.40±7.80  & 83.93±7.33          \\
                           & ViT-L                     &  & \underline{86.00} & \textbf{83.54}±18.37 & \underline{82.85}±21.08 &                      & \underline{88.47}          & \textbf{85.06}±6.79  & \textbf{85.25}±6.61          \\
                           & ViT-H                     &  & 85.54                          & \textbf{83.54}±19.13 & \textbf{82.78}±21.53 &                      & 87.53          & \underline{84.81}±7.65  & 84.78±7.58          \\ \bottomrule
\end{tabular}
\caption{State-of-the-art promptable segmentation model's  performance variation under diverse human-collected bounding boxes for TETRIS and average over all 10 selected general and medical segmentation datasets. Results provided separately for Mobile and Desktop device types. Standard deviation values computed across different users and averaged across the dataset sample. Additionally, we provide IoU, obtained with \textit{tight-bbox} prompt. Best results are in \textbf{bold}, second best are \underline{underlined}.}
\label{tab:user_study4}
\end{table*}

\section{Limitations}

We observe the following limitations of proposed BREPS methodology:

\begin{enumerate}
    \item Small sub-samples in real-user studies, we select 50 images from each of 10 datasets (500 in total), which may not cover all possible instance shapes, occlusions, etc. This limitation is due to the high costs of scaling data collection on real-users. 
    \item We collected data from 2500 users (1250 on desktop and 1250 on mobile devices), which may not cover all possible device resolutions, aspect ratios, user behaviours, etc. We also do not separate users on different devices and dataset instances when creating the CIoU-based Gamma log-PDF regularizer, making one single distribution. We did this to simplify the methodology and obtain more general results across all users, datasets and domains.
    \item Heat maps in the exhaustive search were obtained by reducing the search to only the width and height without shifting the bbox center. Surely we would have obtained more interesting insights if we could have searched all $1024^4$ variants of the bbox prompt, but this is completely unfeasible with our computing resources. However, during a BREPS attack, there are no restrictions on bbox shifts, so the search space is still complete.
    \item We rely on parameter $\lambda$, which defines the tradeoff between prompt realism and attack strength. This parameter was selected in Section~\ref{sec:regularizer} but may not be optimal for each specific instance in the dataset. We believe that even with a fixed tradeoff parameter, it is enough to give an average estimate of the quality and robustness.
    \item Our BREPS attack does not guarantee finding global extrema prompts since it relies on the gradient optimization method (e.g., Adam~\cite{kingma2014adam}). Despite this limitation, it allows us to estimate possible quality boundaries in a feasible time.
    
\end{enumerate}

\newpage
\bibliography{supplement}